\documentclass[conference]{IEEEtran}
\ifCLASSINFOpdf
\else
\fi

\usepackage{enumitem}
\usepackage{amsmath}
\usepackage{float}
\usepackage{amssymb}
\usepackage{graphicx} 
\usepackage{subcaption}
\usepackage{pgfplots}
\usepackage{booktabs}
\usepackage{stfloats}
\usepackage{needspace}
\usepackage{geometry}
\usetikzlibrary{matrix}
\pgfplotsset{compat=1.18}
\usepackage{xcolor}
\usepackage{url}
\definecolor{cICP}{RGB}{220,50,47}
\definecolor{cNF}{RGB}{102,153,255}
\definecolor{cIC}{RGB}{120,190,32}
\usepackage[ruled,vlined,linesnumbered]{algorithm2e}
\SetKwInput{KwInput}{Input}
\SetKwInput{KwOutput}{Output}
\SetKwInput{KwRequire}{Require}
\SetKwInput{KwReturn}{Return}

\begin{document}

\title{\LARGE \bf InvariantCloud: A Globally Invariant, Uniquely Indexed Point Cloud Framework for Robust 6-DoF Tactile Pose Tracking}

\author{%
  \parbox{\linewidth}{\centering
  Pengfei Ye$^{1,\dagger}$, Yuxiang Ma$^{2,\dagger}$, Yi Zhou$^{1}$, Wei Chen$^{3}$, Wenzhen Dong$^{3}$, Molong Duan$^{1,\ast}$\\[0.6em]
  $^{1}$Department of Mechanical and Aerospace Engineering, The Hong Kong University of Science and Technology\\
  $^{2}$Department of Mechanical Engineering, Massachusetts Institute of Technology\\
  $^{3}$Department of Mechanical and Automation Engineering, The Chinese University of Hong Kong\\[0.4em]
  \normalsize
  \textsuperscript{$\dagger$}These authors contributed equally to this work.\quad
  \textsuperscript{$\ast$}Corresponding author.
  }
}

\maketitle

\begin{abstract}
Recent advances in imitation learning and vision–language models highlight the need for high-fidelity tactile perception, with 6-DoF tactile object pose estimation providing a crucial foundation for precise robotic manipulation. We introduce InvariantCloud, a 6-DoF pose estimation framework that leverages the global invariance of surface marker constellations on vision-based tactile sensors. In contrast to recent approaches, our one-shot globally invariant point cloud registration suppresses cumulative drift and overcomes long-standing limitations in accurately estimating yaw (Z-axis) rotation. Experimental verifications show that InvariantCloud achieves superior yaw tracking accuracy and re-localization repeatability compared to existing benchmarks, demonstrating its precision and robustness in long-sequence manipulation tasks.
\end{abstract}

\vspace{2ex}

\begin{IEEEkeywords}
Globally Invariant Point, 6-DoF Pose Estimation, Tactile Sensing
\end{IEEEkeywords}

\IEEEpeerreviewmaketitle

\section{Introduction}
Recent advances in vision–language–action (VLA) models \cite{Fu2024MobileALOHA, Black2024Pi0, Ma2024VLA} have achieved notable grasping and manipulation capabilities. However, without known CAD models and external cameras, high‑precision 6-DoF pose estimation \cite{Gu2021ECPC} of unknown objects remains challenging for these vision‑based systems due to occlusions and incomplete scene understanding. Recent vision–tactile methods adopt dual encoders to handle tactile heterogeneity \cite{Xue2025RDP, Cheng2025OmniVTLA}, yet the reliability of vision-tactile pose estimation is still limited. Touch is often used only as contact or force cues rather than geometric state, which restricts fine‑grained control. To address this limitation and enable more precise manipulation, we target high‑precision, low‑drift 6-DoF object pose tracking using a vision-tactile sensor \cite{GelSightMini, Lin2024_9DTact}.

Many tactile 6-DoF and shape modeling pipelines rely on external cameras and frequent conversions between images and point clouds, introducing cross‑modal errors and cumulative drift \cite{Sodhi2022PatchGraph, Suresh2024NeuralFields, Zhao2023FingerSLAM, Lu2023Tac2Structure}. Normal‑field–based optimization \cite{Huang2025NormalFlow} reduces iterative registration error but often is insufficient to reliably determine yaw (Z‑axis) and to maintain accurate translation. Estimating rotation about the sensor normal is difficult. Torsional limits cause slip that violates the assumed local rigidity. Adding an external RGB camera offers a global view \cite{Zhao2023FingerSLAM, Lu2023Tac2Structure} but increases hardware complexity and sensitivity to illumination. Traditional contour‑based PCA \cite{Wold1987PCA} degrades with blurry or rapidly varying boundaries. Gauss–Newton on the normal field \cite{Huang2025NormalFlow} is capable of recovering yaw only on sufficiently anisotropic surfaces. On locally planar or quasi‑isotropic patches, in‑plane rotation becomes ill-conditioned and the estimation is rendered unreliable.

\begin{figure}[!t]
    \centering
    \includegraphics[width=1.0\linewidth]{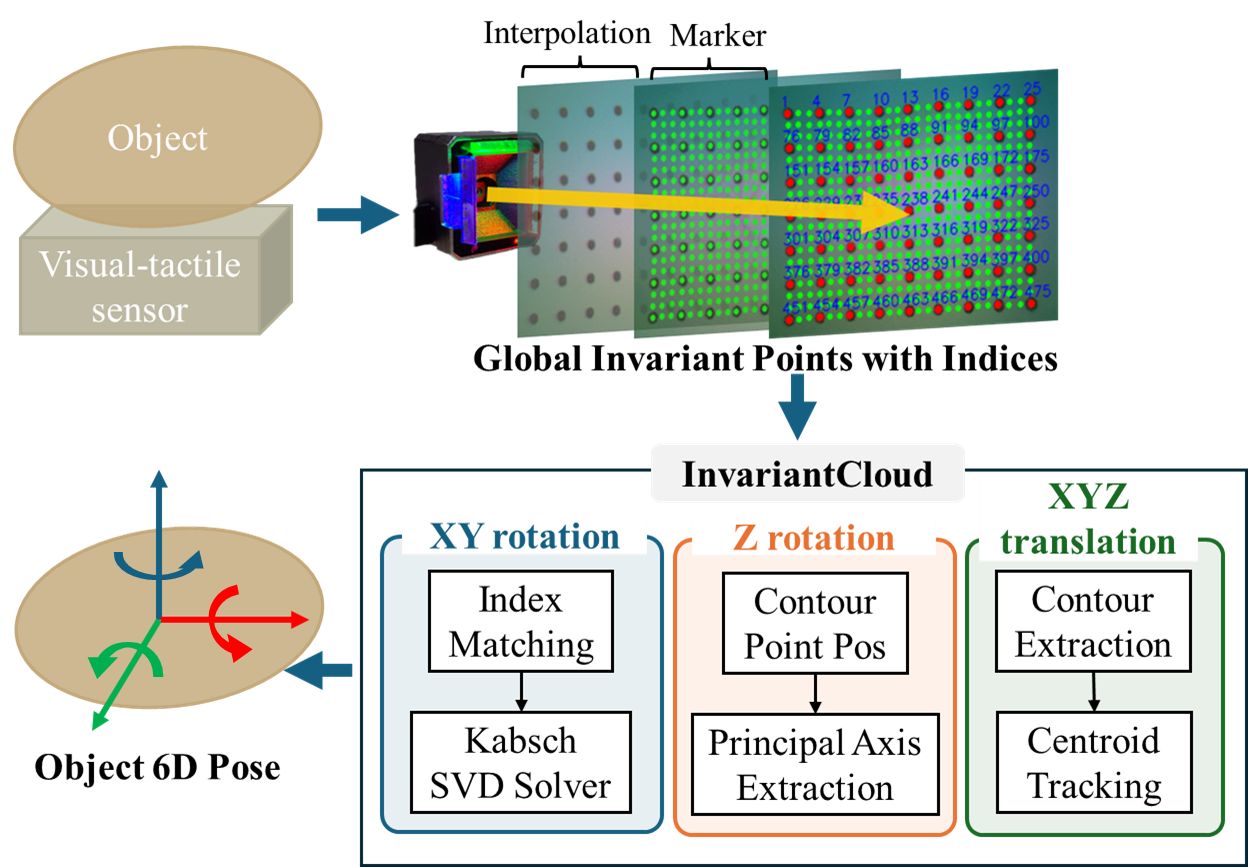}
    \captionsetup{font=footnotesize}
    \caption{A schematic overview of the proposed 6D pose tracking framework, illustrating global invariant point generation, index matching, Kabsch SVD solving for XY rotation, principal axis extraction for Z-axis rotation, and centroid tracking for XYZ translation.}
    \label{fig:schematic_overview}
\end{figure}

Registration, which establishes point‑to‑point correspondences between frames, is a core step in pose estimation but presents significant difficulties with existing approaches. Nearest‑neighbor ICP \cite{Timotee2002FastICP} and frame‑to‑frame Lucas–Kanade optical flow \cite{Baker2019LK} rely on local matches, have higher computational cost, and are prone to mis‑registration that accumulates drift over long sequences. In contrast, when reliable one‑to‑one correspondences are available, the Kabsch SVD solver \cite{Kabsch1978} provides a robust closed‑form solution.

Surface reconstruction from vision–tactile sensing alone has also been explored \cite{Huang2025GelSLAM}. The contact footprint is small, so coverage per touch is limited. When contact breaks, many systems integrate only adjacent frames, and the pose resets on recontact. Traditional large‑area reconstructions often use exteroceptive aids such as motion capture \cite{Huang2025NormalFlow}, depth cameras \cite{Zhao2023FingerSLAM}, or radar. In contrast, some recent approaches attempt to rely solely on tactile sensing for object shape reconstruction. Learning‑based methods map sparse tactile observations to local 3D patches and assemble a global shape with implicit neural representations \cite{Comi2024TouchSDF}, but generalization is difficult and estimates degrade on weakly textured objects. Roller‑based vision–tactile sensors maintain continuous contact \cite{Mirzaee2025GelBelt}, yet their kinematics restrict observable rotations (often small roll/pitch), limiting reconstruction of complex curved surfaces.

To address these gaps, we present InvariantCloud (Fig. 1), a globally invariant, point‑cloud–driven tracking algorithm. We preconstruct in the sensor frame a dense, stable marker point cloud in which each point has a unique ID. During contact, the locally activated subset and its index mapping across frames directly encode the object's rigid‑body motion. This global, uniquely indexable point set removes the need for Lucas–Kanade or ICP search, reduces registration ambiguity, and suppresses cumulative drift. We combine ID‑based one‑to‑one registration with a single‑step Kabsch rotation for in‑plane motion and a PCA principal‑axis strategy for yaw. Experiments demonstrate that the proposed method achieves consistently lower cumulative drift and repeatability errors than existing baselines across all evaluated objects, with particular improvements in yaw estimation accuracy and stability over long sequences.

The main contributions of this paper are as follows:

\begin{enumerate}[label=\arabic*), leftmargin=25pt, labelsep=0.6em, itemsep=0.4em, topsep=0.2em] 
    \item A globally invariant, point‑cloud–based 6-DoF pose estimation framework from tactile sensing with unique ID‑based registration. This design avoids nearest‑neighbor and optical‑flow ambiguity, reduces drift, and enables long‑horizon tracking of unseen objects with low re‑localization error and high repeatability. 
    \item A PCA principal‑axis orientation strategy that leverages the temporal invariance of the global marker layout to accurately recover yaw (Z‑axis), improving rotation estimation success rate and stability over existing methods.
\end{enumerate}

The remainder of this paper is organized as follows. Section II presents the InvariantCloud methodology, detailing the construction of the globally invariant reference point cloud, the Kabsch-based XY rotation estimation, the PCA principal-axis solver for Z-axis rotation, and the centroid-based translation tracking. Additionally, we describe the inter-contact registration framework that enables robust point cloud alignment across temporally separated tactile interactions for long-horizon SLAM applications. Section III provides comprehensive experimental validation across multiple household objects, evaluating static drift, repeatability error, tracking accuracy, and long-term tactile SLAM capabilities. Finally, Section IV discusses the limitations and outlines future research directions.

\section{Method}
\vspace{-2pt}
To enable precise tracking of object poses without requiring prior 3D models, a globally invariant point cloud is leveraged on the surface of a vision-based tactile sensor. A flat, dense reference point cloud is first acquired under a no-contact condition, with a unique identifier (ID) assigned to each point. This spatially stable and uniquely indexable global reference allows direct establishment of inter-frame correspondences via point IDs, effectively eliminating errors in point cloud registration.

\vspace{+5pt}
\setlength{\parindent}{0pt}
{\fontsize{10pt}{15pt}\selectfont \textit{A. Dense Global Reference Point Cloud with Unique IDs}}

\setlength{\parindent}{15pt}
A globally referenced, high-density point cloud with unique identifiers is constructed by first reconstructing a high-accuracy height map generated from object–sensor contact. A sphere of known diameter is pressed into multiple locations on the visuo-tactile sensor surface to collect corresponding images. For each image, RGB color values and normalized pixel coordinates are extracted and used to train a multilayer perceptron (MLP) that predicts the local surface normal’s tilt angle. To improve normal estimation accuracy, pixel-level information from $N_{markers}$ intrinsic fiducial markers embedded in the silicone surface is also included in the network input. After obtaining the gradient field, the height map $H$ is reconstructed by integrating the surface normals using the discrete cosine transform (DCT) solution to the Poisson equation.

After obtaining an accurate height map ($H$), we treat $H$ as a discrete scalar field defined on the integer pixel lattice. The geometric image center and scale normalization, as defined in Eq.~(1), provide a consistent spatial reference for all subsequent computations, where $\rho$ denotes the pixel-to-millimeter scaling factor used to convert image coordinates to physical units. Bilinear interpolation \cite{Kirkland2010Bilinear} (Eq.~(2)) is then used to evaluate a continuous height value at any sub-pixel location. With the geometric image center given in Eq.~(1), each pixel location is uniquely mapped into the global 3D coordinate system via Eq.~(3), achieving both scale normalization and axis alignment between pixel and world frames. Conversely, the inverse transform in Eq.~(4) recovers the corresponding pixel coordinates from any 3D point, thereby establishing a consistent bidirectional mapping between image space and world space.

\begin{align}
&\text{Center: } c_x = \frac{W-1}{2}, \quad c_y = \frac{H-1}{2}, \quad s = \frac{\rho}{1000} \\[4pt]
&\text{Bilinear: } h(x,y) = (1-\alpha)(1-\beta)H[y_0,x_0] \nonumber \\
&\qquad\qquad\qquad\quad + \alpha(1-\beta)H[y_0,x_1] \nonumber \\
&\qquad\qquad\qquad\quad + (1-\alpha)\beta H[y_1,x_0] \nonumber \\
&\qquad\qquad\qquad\quad + \alpha\beta H[y_1,x_1] \\[4pt]
&\text{Pixel}\to\text{World: } P = \left((x-c_x)s, (c_y-y)s, h(x,y)s\right) \\[4pt]
&\text{World}\to\text{Pixel: } (x',y') = \left(\frac{X}{s}+c_x, -\frac{Y}{s}+c_y\right)
\end{align}

Based on this bidirectional pixel-to-3D mapping, we first detect $N_{markers}$ intrinsic fiducial markers arranged in an $R_{orig} \times C_{orig}$ grid on the silicone surface of the sensor. Here, $N_{markers}$ represents the total number of intrinsic fiducial markers embedded in the sensor surface, and $R_{orig}$ and $C_{orig}$ denote the row and column dimensions of their rectangular grid arrangement, respectively. These markers are then densified into a grid using bilinear interpolation, with each interpolated point assigned a globally unique identifier. The detailed algorithmic procedure is provided in Algorithm 1. In principle, this process can generate tens of thousands of points; however, to balance accuracy, robustness, and computational efficiency, we upsample the original $R_{orig} \times C_{orig}$ layout to a denser grid (typically several hundred points), which serves as the global reference for subsequent pose estimation.

\begin{algorithm}[h]
    \setlength{\textfloatsep}{12pt} 
    \caption{Marker Detection and Grid Interpolation}
    \KwInput{RGB image $I$, height map $H$, scale $ppmm$}
    \KwOutput{Interpolated grid pixels $\mathcal{G}$ with global indices}
    \vspace{+5pt}
    
    \textbf{Detection.} Apply an adaptive blob detection until $N_{markers}$ existing points are obtained.
    
    \textbf{Pixel to 3D.} Each marker pixel $m_{i,j}$ is mapped to a 3D world point $P_{i,j}$ using the pixel-to-world formulas (see Eq.~(3)).
    
    \textbf{Interpolation $N_{markers}$$\rightarrow$More.} For every target grid index $(r,c)$ with $0\le r\le m$, $0\le c\le n$:
    \begin{enumerate}
    \item Get interpolation resolutions $a=\lfloor m/(R_{orig}-1)\rfloor$, $b=\lfloor n/(C_{orig}-1)\rfloor$.
    \item Compute $i=\lfloor r/a\rfloor$, $j=\lfloor c/b\rfloor$, $u=(c\bmod b)/b$, $v=(r\bmod a)/a$.
    \item If $u=0$ and $v=0$: $Q_{r,c}=P_{i,j}$.
    \item Else if $v=0$: $Q_{r,c}=(1-u)P_{i,j}+u P_{i,j+1}$.
    \item Else if $u=0$: $Q_{r,c}=(1-v)P_{i,j}+v P_{i+1,j}$.
    \item Else (interior): $Q_{r,c}=(1-u)(1-v)P_{i,j}+u(1-v)P_{i,j+1}+(1-u)vP_{i+1,j}+uvP_{i+1,j+1}$.
    \end{enumerate}
    
    \textbf{Back to Pixels.} Each world point $Q_{r,c}$ is mapped back to pixel coordinates $g_{r,c}$ using the inverse world-to-pixel transform (see Eq.~(4)); assign global index $g = r  (n + 1) + c + 1$.
    
    \textbf{Return} $\mathcal{G}=\{(g,g_{r,c})\}_{r,c}$.
\end{algorithm}

With the mapping and interpolation completed, we store during system initialization a flat, dense, globally invariant reference point set acquired under a no-contact condition (Fig. 2), including each point’s unique identifier and pixel coordinates. This stable point cloud serves as the core reference for subsequent precise object pose tracking.

\begin{figure}[!tb]
    \centering
    \includegraphics[width=0.7\linewidth]{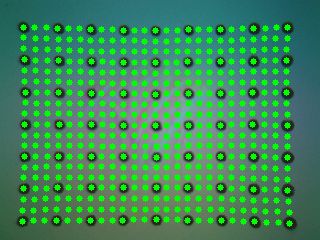}
    \captionsetup{font=footnotesize}
    \caption{Initialization of the dense, globally invariant reference point set under no-contact. The $N_{markers}$ detected fiducial markers are indexed, a structured sparse grid is formed, and marker-guided interpolation yields a dense reference cloud with unique identifiers and stable pixel coordinates used for subsequent pose tracking.}
    \label{fig:initial}
\end{figure}

\vspace{+5pt}
\setlength{\parindent}{0pt}
{\fontsize{10pt}{15pt}\selectfont \textit{B. XY Rotation Estimation with Kabsch Solver}}

\setlength{\parindent}{15pt}
Given the particular treatment required for rotation about the Z-axis, this section focuses only on estimating the object’s in-plane (X/Y-axis) rotation relative to the sensor surface using the globally invariant point cloud; Z-axis rotation will be detailed in the next section. 

Starting from the dense globally referenced point cloud initialized in Fig. 2, each frame is processed by fusing dual thresholds on color difference and height depression, followed by morphological dilation and erosion to produce a stable real-time contact mask $C$. We then obtain the previous and current masks ($C_{prev}$ and $C_{curr}$) and, over the global reference point set, first apply boundary filtering and then index the masks at the points’ integer pixel coordinates to derive Boolean contact subsets. Finally, using Eq.~(3) together with the respective height maps $H_{prev}$ and $H_{curr}$, these 2D contact points are mapped into two corresponding 3D sub–point clouds.

\begin{figure}[!tb]
    \centering
    \includegraphics[width=1.0\linewidth]{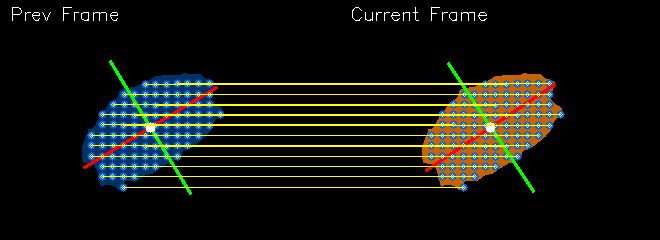}
    \captionsetup{font=footnotesize}
    \caption{One-to-one global ID matching between consecutive contact point sets. Yellow links denote matched indices used directly by the Kabsch algorithm to estimate rotation in a single SVD step.}
    \label{fig:placeholder}
\end{figure}

Since each 3D point in the previous and current contact sub–point clouds carries a unique global identifier, one-to-one correspondences are established directly via the IDs (Fig. 3), and the paired sets are passed to the Kabsch algorithm \cite{Kabsch1978} for a single closed‑form SVD solution of the optimal rigid rotation (Eq.~(7)). We first compute the centroids of both point sets and center the coordinates:

\begin{alignat}{2}
\bar{p}\phantom{_i} &= \frac{1}{N}\sum_{i=1}^N p_i, \quad &\bar{q}\phantom{_i} &= \frac{1}{N}\sum_{i=1}^N q_i \\
\tilde{p}_i &= p_i - \bar{p}, \quad &\tilde{q}_i &= q_i - \bar{q}
\end{alignat}

\setlength{\parindent}{0pt}
The optimal rotation $R^{\star} \in SO(3)$ solves
\begin{equation}
R^{\star} = \arg\min_{R \in SO(3)} \sum_{i=1}^N \|\tilde{p}_i - R\tilde{q}_i\|^2 = \arg\max_{R \in SO(3)} \text{tr}(RH)
\end{equation}
where
\begin{equation}
H = \sum_{i=1}^N \tilde{p}_i \tilde{q}_i^{\top} \in \mathbb{R}^{3\times 3}
\end{equation}

Let the SVD of $H$ be $H = U\Sigma V^{\top}$; then
\begin{equation}
R^{\star} = V \begin{bmatrix} 1 & 0 & 0 \\ 0 & 1 & 0 \\ 0 & 0 & \text{sign}(\det(VU^{\top})) \end{bmatrix} U^{\top}
\end{equation}

\vspace{+5pt}
\setlength{\parindent}{0pt}
{\fontsize{10pt}{15pt}\selectfont \textit{C. PCA principal axis Solver for Z Rotation Estimation}}

\setlength{\parindent}{15pt}
To address the difficulty of reliably estimating rotation about the Z-axis, a PCA principal axis solver driven by a globally invariant dense point cloud is proposed. The method jointly exploits (i) the high stability and drift-free nature of the invariant points under repeated contacts and (ii) the high sensitivity of the contact silhouette to yaw. Except for an ideal sphere, a yaw rotation alters the contact outline and thus the subset of global points it encloses; therefore, in real time, the invariant points within the current silhouette are extracted and their principal axis is computed via PCA (Eq.~(13)). Given the relatively high spatial resolution, even slight contour changes induce a measurable shift in the principal axis, while global invariant indexing guarantees exact reproducibility when the outline returns to a prior pose, preventing cumulative registration drift. Experimental results validate the consistency and robustness of this approach for Z-axis rotation tracking.

\begin{align}
\mathcal{S}_t &= \{p_i \in \mathcal{P} \mid C_t(p_i) = 1\}, \quad K_t = |\mathcal{S}_t|, \\
\mu_t &= \frac{1}{K_t}\sum_{p_i \in \mathcal{S}_t} p_i, \quad \tilde{p}_i = p_i - \mu_t, \\
M_t &= \frac{1}{K_t}\sum_{p_i \in \mathcal{S}_t} \tilde{p}_i \tilde{p}_i^{\top} = U_t\Sigma_t U_t^{\top}, \\
a_t &= U_t(:,1), \quad \theta_t = \text{atan2}(a_{t,y}, a_{t,x}).
\end{align}

In the above equations, $\mathcal{P}$ represents the globally invariant dense point set containing all registered 2D points, while $C_t$ denotes the binary contact mask at frame $t$ that maps points in $\mathcal{P}$ to contact states via sampling. The contact subset $\mathcal{S}_t$ contains all points from $\mathcal{P}$ that are in contact at frame $t$, with cardinality $K_t$. Each point $p_i \in \mathbb{R}^2$ represents the 2D coordinates of a selected invariant point in planar projection. The covariance matrix $M_t \in \mathbb{R}^{2\times 2}$ captures the spatial distribution of contact points, and $a_t$ denotes the principal axis obtained from the first column of $U_t$.

\begin{figure}[h]
\centering
\setlength{\tabcolsep}{4pt}
\renewcommand{\arraystretch}{1.0}
\scalebox{1.0}{%
\begin{tabular}{cc}
\textbf{NormalFlow \cite{Huang2025NormalFlow}} & \textbf{InvariantCloud (Ours)} \\

\subcaptionbox{Initial\label{fig:ztrack:a}}{%
  \includegraphics[width=0.47\columnwidth]{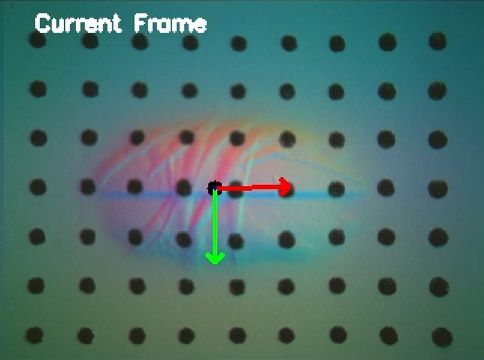}} &
\subcaptionbox{Initial\label{fig:ztrack:b}}{%
  \includegraphics[width=0.47\columnwidth]{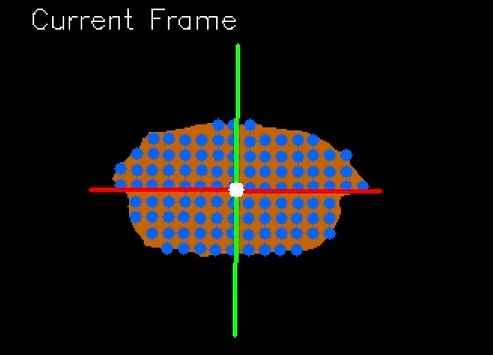}} \\

\subcaptionbox{Early drift\label{fig:ztrack:c}}{%
  \includegraphics[width=0.47\columnwidth]{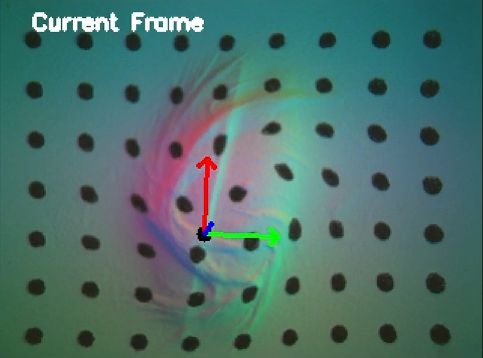}} &
\subcaptionbox{Early drift\label{fig:ztrack:d}}{%
  \includegraphics[width=0.47\columnwidth]{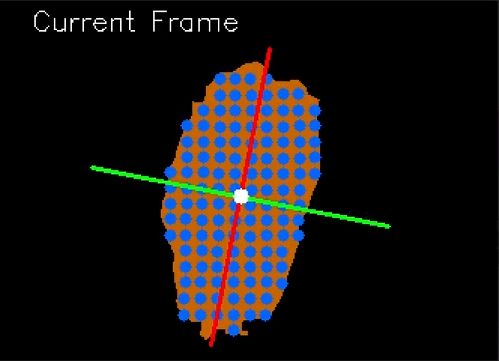}} \\

\subcaptionbox{Severe drift\label{fig:ztrack:e}}{%
  \includegraphics[width=0.47\columnwidth]{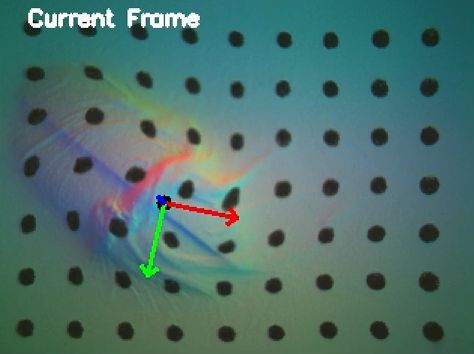}} &
\subcaptionbox{Severe drift\label{fig:ztrack:f}}{%
  \includegraphics[width=0.47\columnwidth]{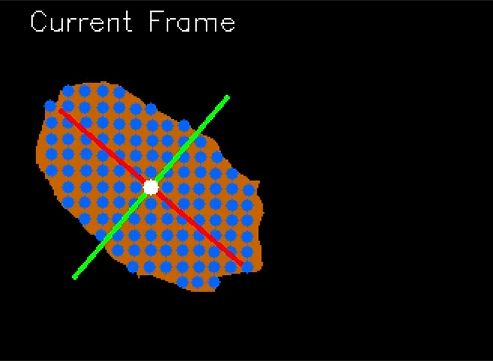}} \\

\subcaptionbox{Return / repeatability \label{fig:ztrack:g}}{%
  \includegraphics[width=0.47\columnwidth]{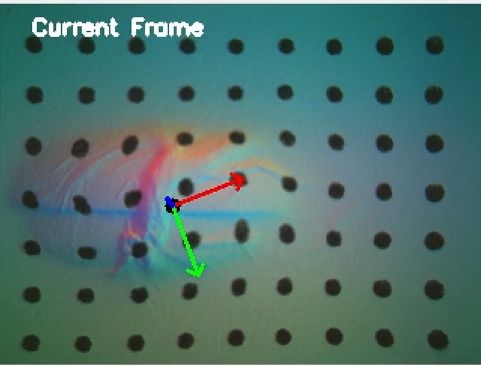}} &
\subcaptionbox{Return / repeatability \label{fig:ztrack:h}}{%
  \includegraphics[width=0.47\columnwidth]{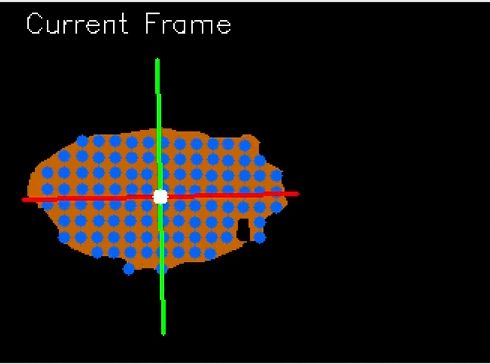}} \\
\end{tabular}}
\captionsetup{font=footnotesize}
\caption{Comparison of Z-axis rotation tracking across four stages: initial alignment, onset of drift, severe drift, and return-to-start repeatability. During Z-axis rotation the surface markers visibly shear and slip relative to the contacting object, whereas our globally invariant point cloud remains spatially uniform and temporally stable throughout, providing a drift‑free geometric reference.}
\label{fig:ztrack}
\end{figure}

\vspace{+5pt}
\setlength{\parindent}{0pt}
{\fontsize{10pt}{15pt}\selectfont \textit{D. Contact Centroid-Based XYZ Translation Estimation}}

\setlength{\parindent}{15pt}
To track the object’s 3D (XYZ) translation on the sensor surface, we take the geometric centroid of the current contact subset within the globally invariant point cloud as the sole tracking target. This strategy persistently anchors the contact region centroid, reducing failure cases and jitter over long sequences. The globally invariant indexing also suppresses inter-frame drift, keeping cumulative error minimal. 

\begin{algorithm}[h]
\setlength{\textfloatsep}{12pt}
\caption{Contact Centroid Estimation}
\KwInput{Binary mask $C_{\text{tar}}$}
\KwOutput{3-D centroid $c_t$}

\textbf{Close.} $C_{\text{closed}} \leftarrow \text{CLOSE}(C_{\text{tar}}, \text{ellipse}, 7, 2)$.

\textbf{Largest contour.} $\gamma \leftarrow \arg\max_{\gamma_i\in\text{Contours}(C_{\text{closed}})} \text{Area}(\gamma_i)$.

\textbf{Centroid.} $(c_x,c_y) \leftarrow \frac{1}{|\gamma|} \sum_{(x,y)\in\gamma}(x,y)$.

\textbf{World lift.} $c_t \leftarrow \pi^{-1}\bigl([c_x,c_y]\bigr)$.

\textbf{Return} $c_t$.
\end{algorithm}

Implementation-wise, we first apply a morphological closing to the contact mask to obtain a stable dominant contour, then compute its area-moment centroid and lift it to 3D; frame-to-frame centroid differences yield XY translation, while the Z component is estimated from the contact region height. The complete procedure is summarized in Algorithm 2.

\vspace{+5pt}
\setlength{\parindent}{0pt}
{\fontsize{10pt}{15pt}\selectfont \textit{E. Inter-Contact Registration for Long-Horizon Tactile SLAM}}

\setlength{\parindent}{15pt}
Reliable inter-contact registration is fundamental to achieving long-horizon tactile SLAM \cite{Huang2025GelSLAM, Bauza2023Tac2Pose}, particularly when reconstructing object surfaces that are significantly larger than the sensor's contact footprint. Building upon the robust 6-DoF pose estimation framework described above, InvariantCloud extends its capabilities to enable such cross-contact registration through its globally invariant point cloud representation.

The key innovation lies in leveraging the absolute Z-axis orientation provided by our PCA principal-axis method. Unlike frame-to-frame tracking approaches that only provide relative pose transformations between adjacent contacts, PCA provides an absolute principal axis for every contact event, enabling robust registration across temporally separated contacts. This absolute orientation reference allows reliable pairing of point clouds even when contact is interrupted and resumed, which is essential for comprehensive surface reconstruction using vision-based tactile sensors.

Beyond pose estimation, the quality of point cloud sampling also plays a crucial role in reconstruction fidelity. Unlike methods that randomly generate points within the contact region, our approach leverages the globally invariant, uniquely indexed reference cloud to sample points directly from the interpolation grid. This strategy improves robustness to minor hand tremors and reduces spurious outliers. During registration, correspondences are anchored to these persistent indices, preserving point identity across contacts and enabling long-horizon accumulation without drift.

The inter-contact registration workflow (Fig. 5) proceeds as follows: the rotational difference about the Z-axis between successive point clouds is computed using their respective principal axes, followed by XY-plane alignment and geometric centering. ICP is then applied to refine the registration, with overlap ratio computed to assess matching reliability.

\begin{figure}[!tb]
    \centering
    \includegraphics[width=0.48\textwidth]{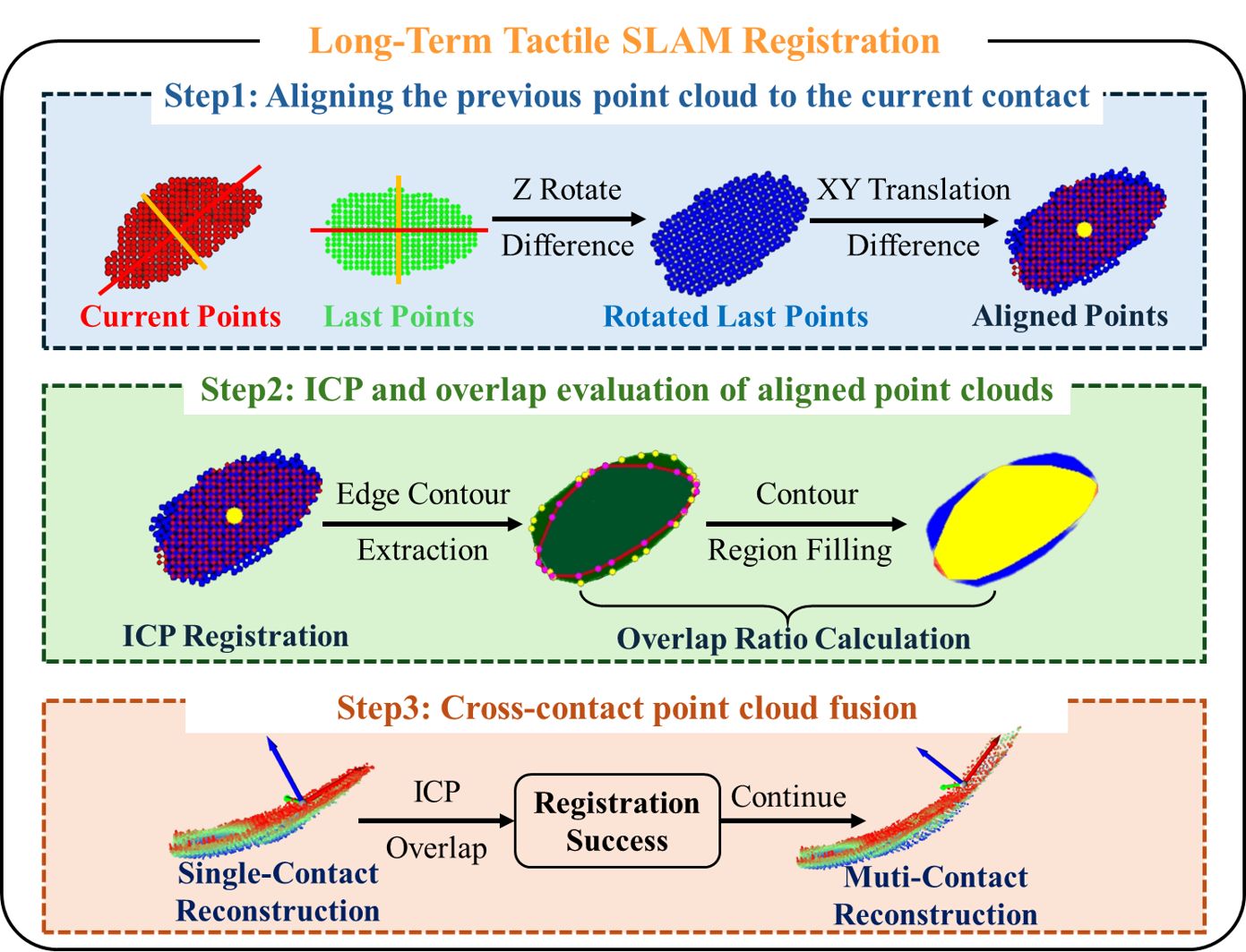}
    \captionsetup{font=footnotesize}
    \caption{Workflow of point cloud registration between successive contacts. In the first two steps, the two point clouds cannot be merged due to insufficient overlap or poor ICP alignment. The third step illustrates successful registration and integration across contacts.}
    \label{fig:your_label}
\end{figure}

\section{Experiments and Results}
This section evaluates the performance of the proposed InvariantCloud method for full 6D pose tracking of several common household objects (see Fig. 6). All experiments were conducted using the GelSight Mini visual-tactile sensor \cite{GelSightMini} (resolution: 320×240, frame rate: $\approx$ 25 Hz), with marker points upsampled to a 19×25 grid (475 points) as the global reference. Two baselines are compared: (1) Lucas-Kanade (LK) optical flow with nearest-neighbor ICP; and (2) NormalFlow \cite{Huang2025NormalFlow} via Gauss-Newton optimization on surface normals. All coordinate systems follow the Z-X-Y Euler right-handed convention.

Performance is quantified by three metrics targeting long-term stability, return consistency, and instantaneous accuracy, respectively:
\begin{enumerate}[label=\arabic*), leftmargin=25pt, 
labelsep=0.6em, itemsep=0.4em, topsep=0.2em]
  \item Static cumulative drift: accumulated pose deviation 
  under stationary contact, averaged over five sequences 
  per method to mitigate hand tremors.
  \item Repeatability (return) error: pose error after 
  displacing and returning a representative household 
  object to its initial contact pose, compared against 
  the expected zero pose.
  \item Tracking accuracy: deviation between estimated and 
  ground-truth pose during single-axis motion across a 
  clearly distinguishable range.
\end{enumerate}

Experiments~A--C each target one of the above metrics; Experiment~A additionally provides the contour matching baseline used to verify return poses in Experiment~B. Experiment~D further validates long-term sequential surface modeling using the absolute Z-axis rotation unique to InvariantCloud, with grid density increased to 31×41 for enhanced fidelity.

\renewcommand{\arraystretch}{1.2}
\begin{table}[h]
\centering
\caption{Static Cumulative MAE for All Objects During One-Minute Stationary Contact}
\begin{tabular}{lcccccc}
\hline
 & \multicolumn{3}{c}{Translation (mm)} & 
   \multicolumn{3}{c}{Rotation (°)} \\
\cmidrule(lr){2-4}\cmidrule(lr){5-7}
Method & $\Delta x$ & $\Delta y$ & $\Delta z$ & 
$\Delta\theta_x$ & $\Delta\theta_y$ & $\Delta\theta_z$ \\
\hline
\textbf{InvariantCloud} & \textbf{0.22} & \textbf{0.14} & \textbf{0.11} & \textbf{0.62} & \textbf{0.69} & \textbf{0.84} \\
NormalFlow & 0.41 & 0.38 & 0.28 & 1.42 & 0.97 & 0.78 \\
ICP & 2.52 & 3.14 & 2.29 & 10.89 & 8.15 & 7.19 \\
\hline
\end{tabular}
\end{table}

\vspace{+5pt}
\setlength{\parindent}{0pt}
{\fontsize{10pt}{15pt}\selectfont \textit{A. Static Cumulative Drift Measurement}}

\setlength{\parindent}{15pt}
In this experiment, we focus on evaluating the cumulative error measured by three algorithms while the object remains stationary for one minute. To minimize instability caused by hand-held operation, we compare the similarity between the real-time contact region contour acquired by the sensor and that of the initial frame, ensuring the contact area remains consistent throughout the test. As shown in Table I, both NormalFlow and the proposed InvariantCloud method exhibit excellent static error performance over long stationary periods, whereas the conventional LK+ICP approach gradually accumulates static drift, resulting in increasingly unreliable measurements.

\begin{figure*}[ht]
\centering
% =================== 全局图例 ===================
\begin{flushleft}
\hspace{40pt}
\scriptsize
\fbox{\begin{tabular}{@{\hspace{4pt}}c@{\hspace{2pt}}c@{\hspace{8pt}}c@{\hspace{2pt}}c@{\hspace{8pt}}c@{\hspace{2pt}}c@{\hspace{4pt}}}
\tikz\draw[draw=cICP,fill=cICP!55,line width=0.3pt] (0,0) rectangle (8pt,6pt); & ICP &
\tikz\draw[draw=cNF,fill=cNF!55,line width=0.3pt] (0,0) rectangle (8pt,6pt); & NormalFlow &
\tikz\draw[draw=cIC,fill=cIC!55,line width=0.3pt] (0,0) rectangle (8pt,6pt); & InvariantCloud (Ours)
\end{tabular}}
\end{flushleft}

\vspace{8pt}

\setlength{\tabcolsep}{1pt}
% =================== 第一行 ===================
\begin{tabular}{c c c c c c c}
\begin{tabular}{c}
  \includegraphics[width=0.09\textwidth]{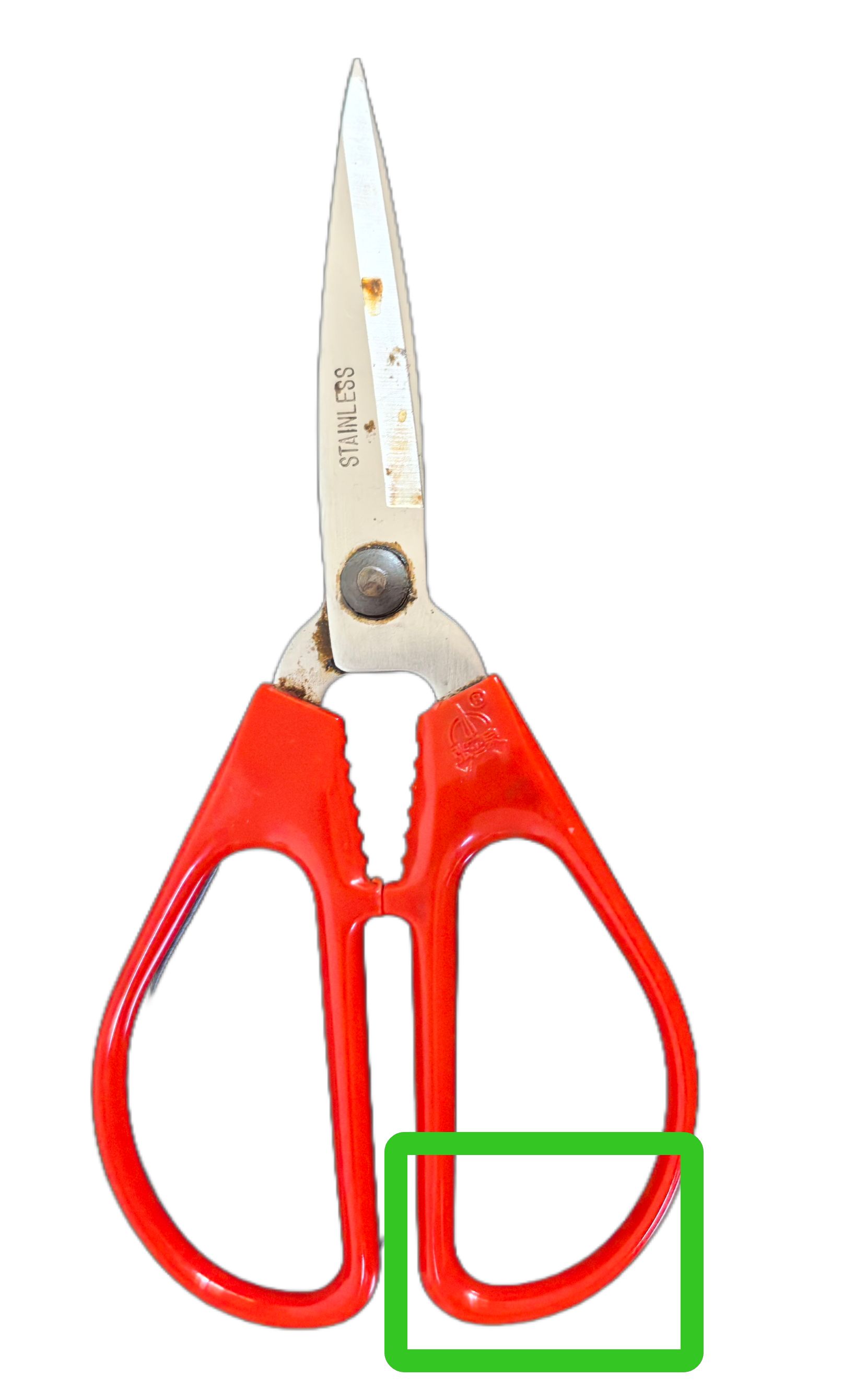} \\
  \includegraphics[width=0.09\textwidth]{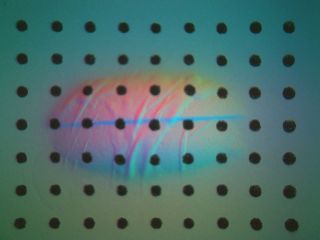}
\end{tabular} &
\begin{tikzpicture}[baseline, yshift=-24pt]
\begin{axis}[width=0.20\textwidth, height=0.20\textwidth, ybar, bar width=5pt, ymin=0, ymax=10, ylabel={Trans. Err (mm)}, ytick={0,5,10},symbolic x coords={X,Y,Z}, xtick=data, enlarge x limits=0.20, ymajorgrids, grid style={gray!30}, tick label style={font=\scriptsize}, label style={font=\scriptsize}, every axis plot/.append style={line width=0.3pt}]
\addplot+[draw=cICP,fill=cICP!55] coordinates {(X,4.4)(Y,3.8)(Z,2.1)};
\addplot+[draw=cNF, fill=cNF!55] coordinates {(X,1.8)(Y,1.9)(Z,1.2)};
\addplot+[draw=cIC, fill=cIC!55] coordinates {(X,1.2)(Y,1.3)(Z,0.7)};
\end{axis}
\end{tikzpicture} &
\begin{tikzpicture}[baseline, yshift=-24pt]
\begin{axis}[
    width=0.20\textwidth,
    height=0.20\textwidth,
    ybar,
    bar width=5pt,
    ymin=0, ymax=20,
    ylabel={Rot. Err ($^\circ$)},
    ytick={0,10,20}, % <-- 这里
    symbolic x coords={$\theta_x$,$\theta_y$,$\theta_z$},
    xtick=data,
    enlarge x limits=0.20,
    ymajorgrids,
    grid style={gray!30},
    tick label style={font=\scriptsize},
    label style={font=\scriptsize},
    every axis plot/.append style={line width=0.3pt}
]
\addplot+[draw=cICP,fill=cICP!55] coordinates {($\theta_x$,4.9)($\theta_y$,3.7)($\theta_z$,6.5)};
\addplot+[draw=cNF, fill=cNF!55] coordinates {($\theta_x$,2.8)($\theta_y$,2.4)($\theta_z$,2.1)};
\addplot+[draw=cIC, fill=cIC!55] coordinates {($\theta_x$,1.4)($\theta_y$,1.2)($\theta_z$,1.3)};
\end{axis}
\end{tikzpicture} &
\quad &
\begin{tabular}{c}
  \includegraphics[width=0.09\textwidth]{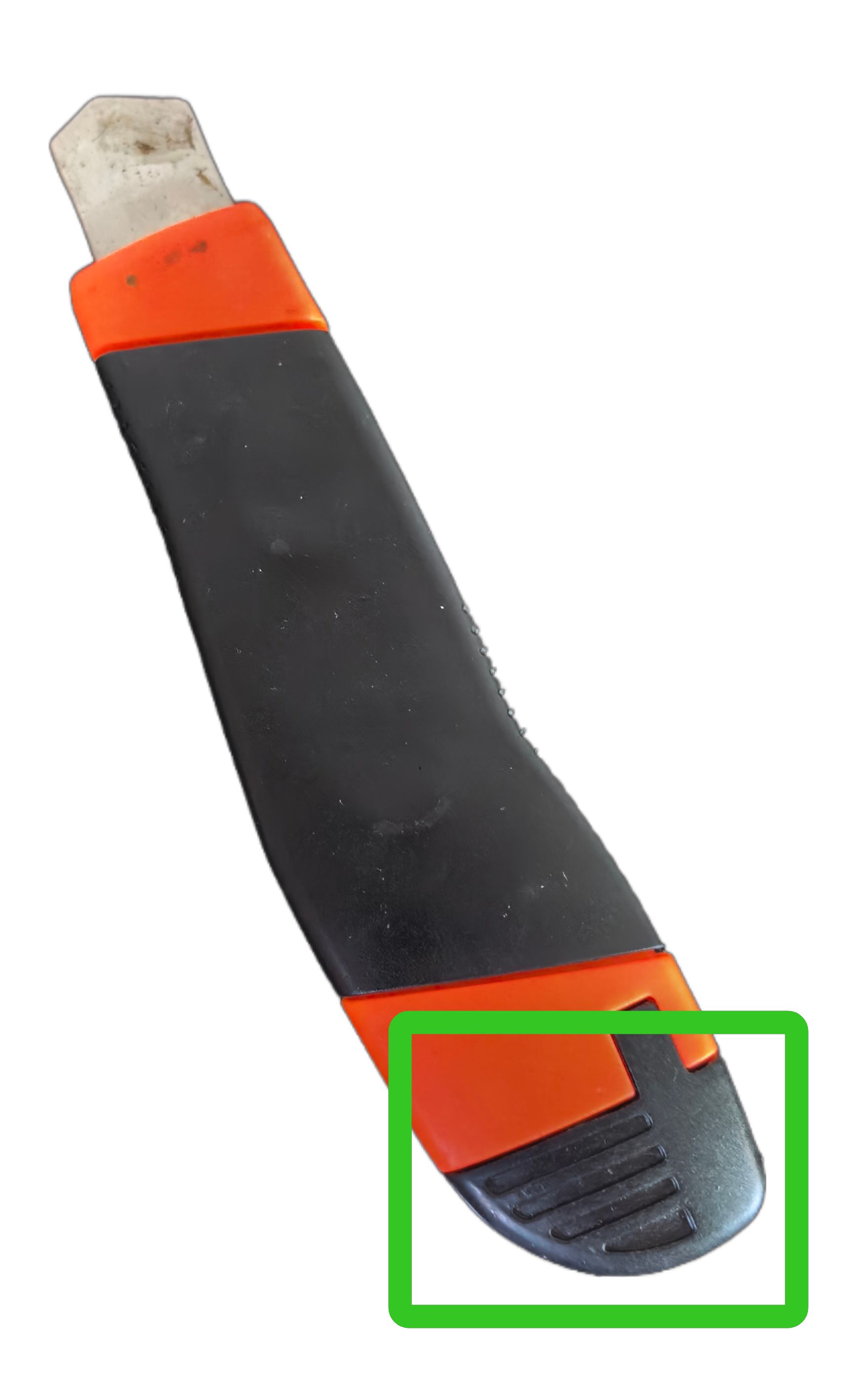} \\
  \includegraphics[width=0.09\textwidth]{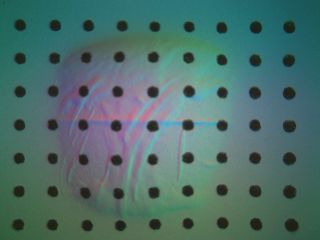}
\end{tabular} &
\begin{tikzpicture}[baseline, yshift=-24pt]
\begin{axis}[width=0.20\textwidth, height=0.20\textwidth, ybar, bar width=5pt, ymin=0, ymax=10, ylabel={Trans. Err (mm)}, ytick={0,5,10}, symbolic x coords={X,Y,Z}, xtick=data, enlarge x limits=0.20, ymajorgrids, grid style={gray!30}, tick label style={font=\scriptsize}, label style={font=\scriptsize}, every axis plot/.append style={line width=0.3pt}]
\addplot+[draw=cICP,fill=cICP!55] coordinates {(X,5.9)(Y,4.5)(Z,2.5)};
\addplot+[draw=cNF, fill=cNF!55] coordinates {(X,3.5)(Y,2.35)(Z,1.8)};
\addplot+[draw=cIC, fill=cIC!55] coordinates {(X,1.4)(Y,1.2)(Z,0.9)};
\end{axis}
\end{tikzpicture} &
\begin{tikzpicture}[baseline, yshift=-24pt]
\begin{axis}[
    width=0.20\textwidth,
    height=0.20\textwidth,
    ybar,
    bar width=5pt,
    ymin=0, ymax=20,
    ylabel={Rot. Err ($^\circ$)},
    ytick={0,10,20}, % <-- 这里
    symbolic x coords={$\theta_x$,$\theta_y$,$\theta_z$},
    xtick=data,
    enlarge x limits=0.20,
    ymajorgrids,
    grid style={gray!30},
    tick label style={font=\scriptsize},
    label style={font=\scriptsize},
    every axis plot/.append style={line width=0.3pt}
]
\addplot+[draw=cICP,fill=cICP!55] coordinates {($\theta_x$,5.5)($\theta_y$,4.6)($\theta_z$,9.8)};
\addplot+[draw=cNF, fill=cNF!55] coordinates {($\theta_x$,2.7)($\theta_y$,2.55)($\theta_z$,6.0)};
\addplot+[draw=cIC, fill=cIC!55] coordinates {($\theta_x$,1.62)($\theta_y$,1.45)($\theta_z$,1.51)};
\end{axis}
\end{tikzpicture}
\\[-2pt]
\scriptsize Scissors & & & \hspace{10pt} & \scriptsize Utility knife & &
\end{tabular}

\vspace{8pt}

% =================== 第二行 ===================
\begin{tabular}{c c c c c c c}
\begin{tabular}{c}
  \includegraphics[width=0.09\textwidth]{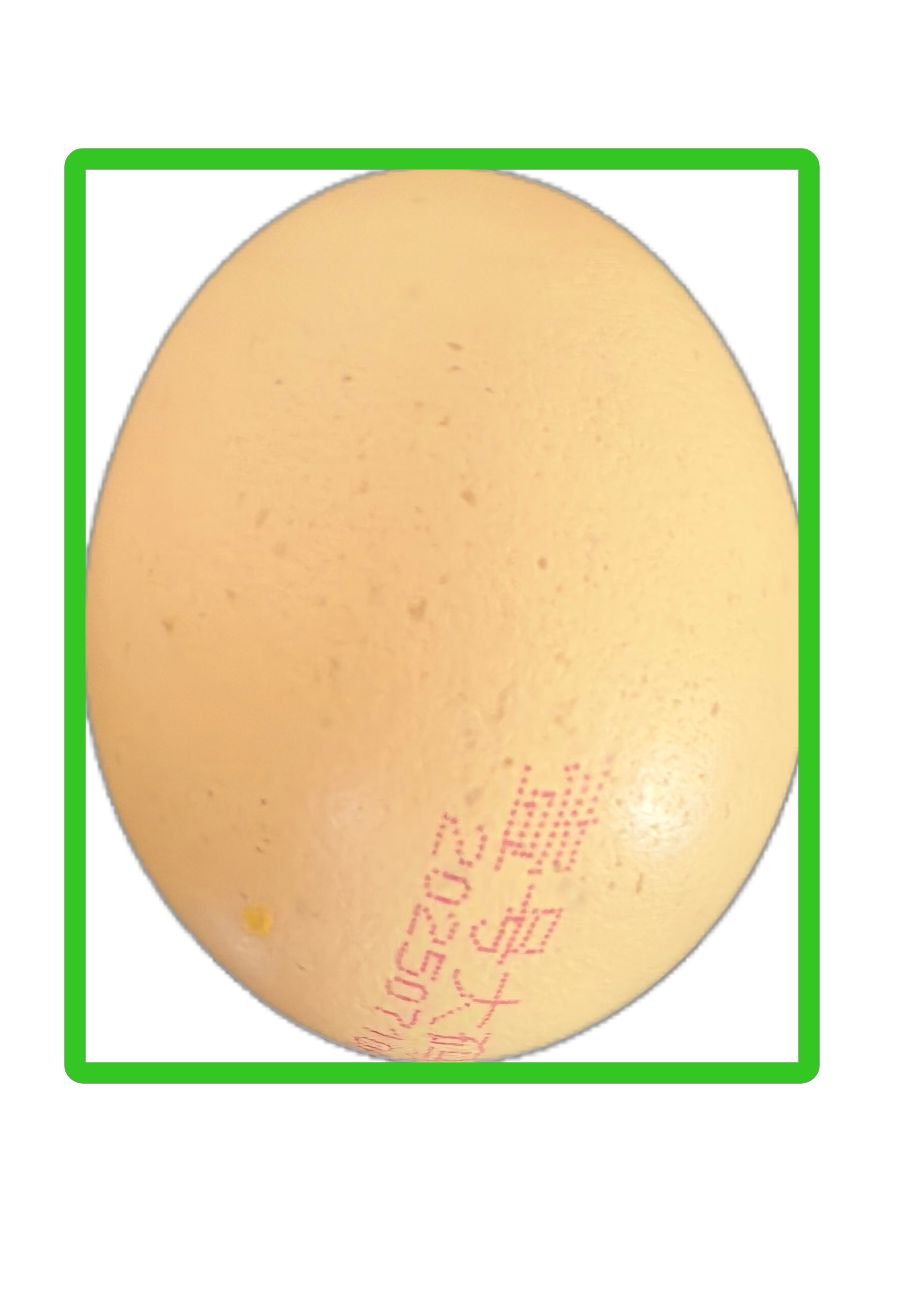} \\
  \includegraphics[width=0.09\textwidth]{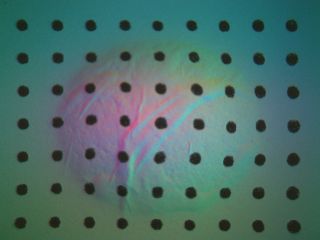}
\end{tabular} &
\begin{tikzpicture}[baseline, yshift=-24pt]
\begin{axis}[width=0.20\textwidth, height=0.20\textwidth, ybar, bar width=5pt, ymin=0, ymax=10, ylabel={Trans. Err (mm)}, ytick={0,5,10}, symbolic x coords={X,Y,Z}, xtick=data, enlarge x limits=0.20, ymajorgrids, grid style={gray!30}, tick label style={font=\scriptsize}, label style={font=\scriptsize}, every axis plot/.append style={line width=0.3pt}]
\addplot+[draw=cICP,fill=cICP!55] coordinates {(X,7.2)(Y,8.7)(Z,2.4)};
\addplot+[draw=cNF, fill=cNF!55] coordinates {(X,4.4)(Y,5.3)(Z,2.6)};
\addplot+[draw=cIC, fill=cIC!55] coordinates {(X,2.3)(Y,2.8)(Z,1.5)};
\end{axis}
\end{tikzpicture} &
\begin{tikzpicture}[baseline, yshift=-24pt]
\begin{axis}[
    width=0.20\textwidth,
    height=0.20\textwidth,
    ybar,
    bar width=5pt,
    ymin=0, ymax=20,
    ylabel={Rot. Err ($^\circ$)},
    ytick={0,10,20}, % <-- 这里
    symbolic x coords={$\theta_x$,$\theta_y$,$\theta_z$},
    xtick=data,
    enlarge x limits=0.20,
    ymajorgrids,
    grid style={gray!30},
    tick label style={font=\scriptsize},
    label style={font=\scriptsize},
    every axis plot/.append style={line width=0.3pt}
]
\addplot+[draw=cICP,fill=cICP!55] coordinates {($\theta_x$,7.5)($\theta_y$,6.8)($\theta_z$,18.2)};
\addplot+[draw=cNF, fill=cNF!55] coordinates {($\theta_x$,4.5)($\theta_y$,3.4)($\theta_z$,14.8)};
\addplot+[draw=cIC, fill=cIC!55] coordinates {($\theta_x$,2.4)($\theta_y$,2.3)($\theta_z$,2.7)};
\end{axis}
\end{tikzpicture} &
\quad &
\begin{tabular}{c}
  \includegraphics[width=0.09\textwidth]{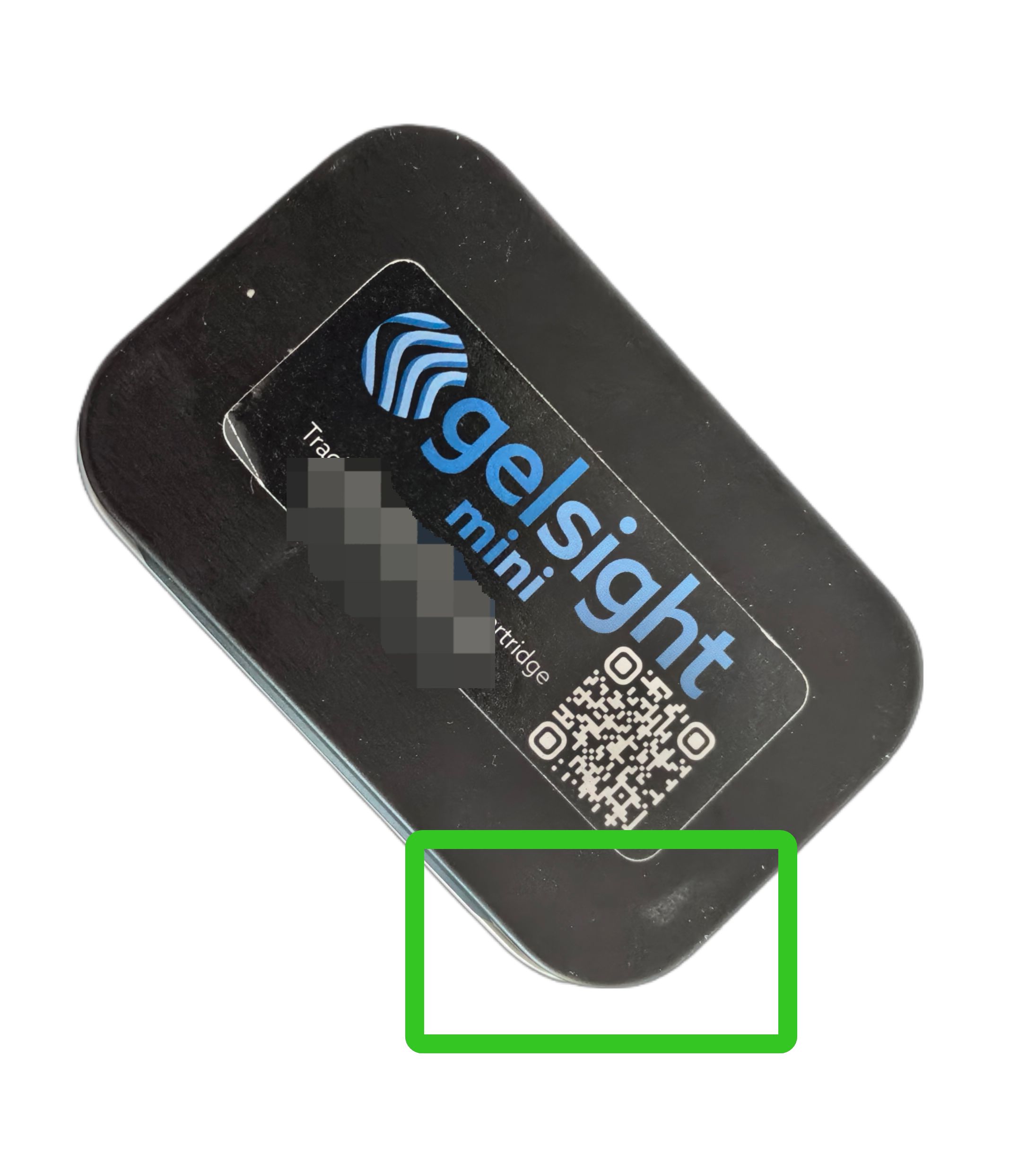} \\
  \includegraphics[width=0.09\textwidth]{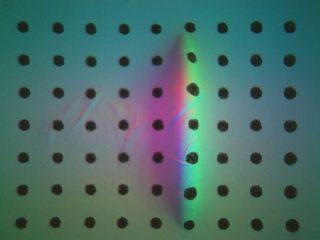}
\end{tabular} &
\begin{tikzpicture}[baseline, yshift=-24pt]
\begin{axis}[width=0.20\textwidth, height=0.20\textwidth, ybar, bar width=5pt, ymin=0, ymax=10, ylabel={Trans. Err (mm)}, ytick={0,5,10}, symbolic x coords={X,Y,Z}, xtick=data, enlarge x limits=0.20, ymajorgrids, grid style={gray!30}, tick label style={font=\scriptsize}, label style={font=\scriptsize}, every axis plot/.append style={line width=0.3pt}]
\addplot+[draw=cICP,fill=cICP!55] coordinates {(X,3.2)(Y,2.7)(Z,1.9)};
\addplot+[draw=cNF, fill=cNF!55] coordinates {(X,1.6)(Y,1.4)(Z,1.1)};
\addplot+[draw=cIC, fill=cIC!55] coordinates {(X,0.9)(Y,0.8)(Z,0.6)};
\end{axis}
\end{tikzpicture} &
\begin{tikzpicture}[baseline, yshift=-24pt]
\begin{axis}[
    width=0.20\textwidth,
    height=0.20\textwidth,
    ybar,
    bar width=5pt,
    ymin=0, ymax=20,
    ylabel={Rot. Err ($^\circ$)},
    ytick={0,10,20}, % <-- 这里
    symbolic x coords={$\theta_x$,$\theta_y$,$\theta_z$},
    xtick=data,
    enlarge x limits=0.20,
    ymajorgrids,
    grid style={gray!30},
    tick label style={font=\scriptsize},
    label style={font=\scriptsize},
    every axis plot/.append style={line width=0.3pt}
]
\addplot+[draw=cICP,fill=cICP!55] coordinates {($\theta_x$,4.8)($\theta_y$,2.9)($\theta_z$,5.1)};
\addplot+[draw=cNF, fill=cNF!55] coordinates {($\theta_x$,1.7)($\theta_y$,1.5)($\theta_z$,1.9)};
\addplot+[draw=cIC, fill=cIC!55] coordinates {($\theta_x$,0.9)($\theta_y$,1.1)($\theta_z$,1.0)};
\end{axis}
\end{tikzpicture}
\\[-2pt]
\scriptsize Egg & & & \hspace{10pt} & \scriptsize Sensor box & &
\end{tabular}
\captionsetup{font=footnotesize}
\caption{This figure presents the pose errors computed by three methods for four common daily objects after single-axis rotation or translation and subsequent return to a configuration nearly overlapping with the sensor’s initial frame. For each method and object, five independent trials were performed and averaged to minimize the influence of random factors and enhance the reliability of the results. The green boxed regions indicate the actual contact areas between the object and the sensor.}
\label{fig:bars_two_rows_four_objects}
\end{figure*}
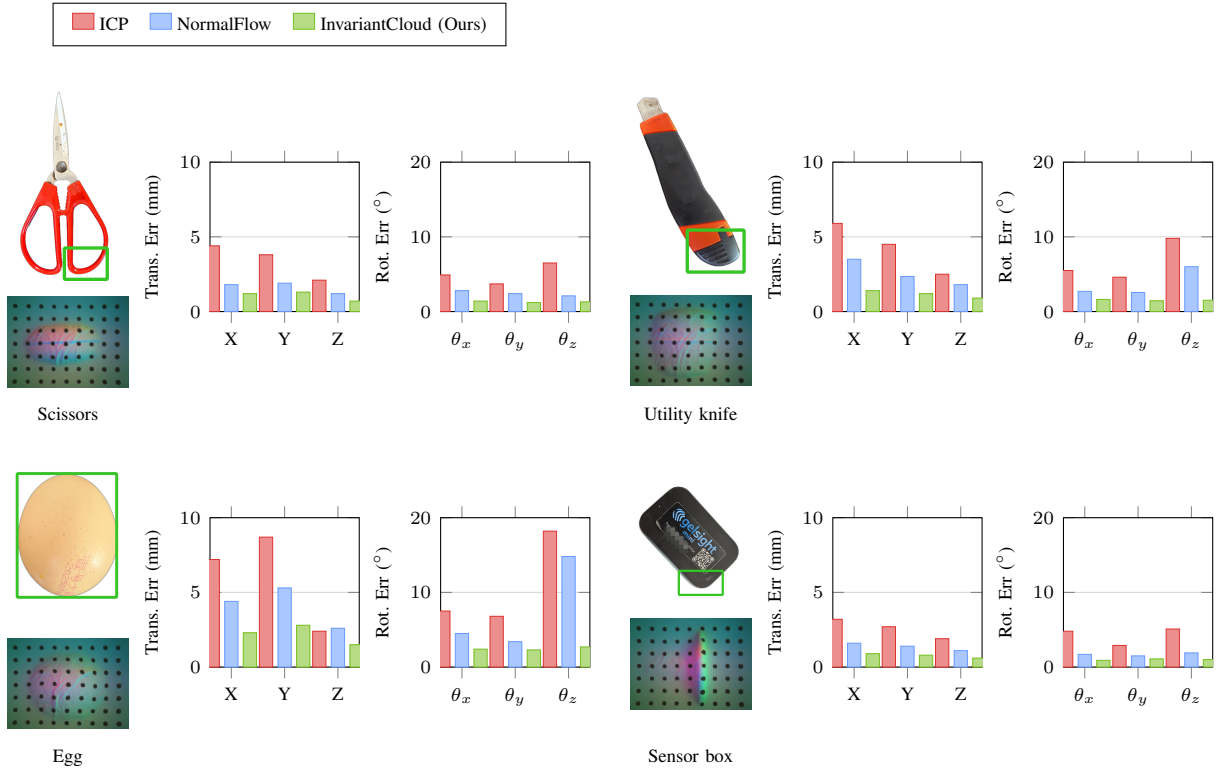

\needspace{4\baselineskip}
\vspace{+5pt}
\setlength{\parindent}{0pt}
{\fontsize{10pt}{15pt}\selectfont \textit{B. Repeatability (Return) Error Evaluation}}

\setlength{\parindent}{15pt}
High repeatability in pose estimation is essential for many upstream tasks, such as tactile SLAM \cite{Zhao2023FingerSLAM} and force-position coordinated manipulation \cite{Zhou2025Bricklaying}, where consistent localization accuracy is critical to avoid significant modeling errors in overlapping regions. In this section, we evaluate four objects by randomly rotating or translating them (while maintaining contact with the sensor), and then returning them to a contact region closely matching the initial frame, as verified by the contour matching algorithm described in Experiment A. The pose output by each algorithm is compared to the expected zero pose to quantify the repeatability error. To ensure reliability, each experiment is repeated five times per object and method. The detailed results are presented in Fig. 6.

It is evident that ICP fails to deliver competitive performance, exhibiting significant errors even under static conditions and thus proving unsuitable for dynamic tests. NormalFlow maintains reasonable accuracy for rotations about the X and Y axes in this experiment; however, its repeatability error for Z-axis rotation—particularly for objects like eggs with elliptical contact surfaces—can approach 20 degrees, and its performance in XYZ displacement is also suboptimal. This phenomenon is further illustrated in Fig. 4. In contrast, the proposed InvariantCloud method consistently achieves the lowest repeatability errors among all compared methods for all four objects, outperforming NormalFlow and ICP by a significant margin across all degrees of freedom.

\vspace{+5pt}
\setlength{\parindent}{0pt}
{\fontsize{10pt}{15pt}\selectfont \textit{C. Tracking Accuracy Analysis}}

\setlength{\parindent}{15pt}
Accurate estimation of object rotation and displacement on the sensor surface is essential for precise manipulation tasks, such as visual-language-action (VLA), particularly when a robotic arm operates in confined or dark environments where visual feedback from eye-in-hand or head-mounted cameras is unavailable. In these scenarios, tactile sensing becomes the sole source for pose estimation. Conventional ICP methods (see Fig. 7) rely on tracking corner points on the sensor surface; however, during Z-axis rotation, objects frequently slip, rendering ICP ineffective for tracking rotational motion about the Z-axis. Building on previous results, and given that NormalFlow already performs well for X and Y axis rotations, this experiment focuses on comparing the Z-axis rotational tracking performance of NormalFlow and the proposed InvariantCloud method. The experimental setup is illustrated in Fig. 8, where three fluorescent markers are attached to each object’s surface to enable motion capture cameras to track the trajectory and ensure the rotation angle reaches 90 degrees.

\begin{figure}[!tb]
    \centering
    \includegraphics[width=0.19\textwidth]{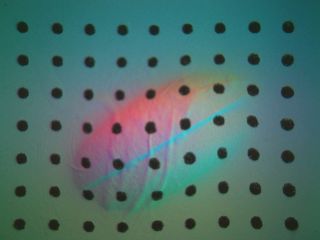}
    \hspace{8pt}
    \includegraphics[width=0.19\textwidth]{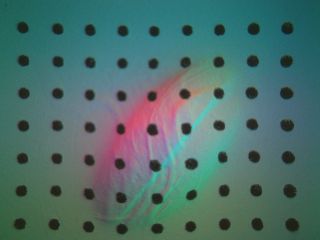}
    \captionsetup{font=footnotesize}
    \caption{Illustration of Z-axis slip: object rotation causes relative slip on the sensor surface, leaving fiducial markers nearly stationary and rendering ICP ineffective for Z-axis rotation estimation.}
    \label{fig:icp_slip_demo}
\end{figure}

\begin{figure}[!h]
\centering
\setlength{\tabcolsep}{2pt} % 列间距更小
\renewcommand{\arraystretch}{1.0}
\scalebox{1.0}{%
\begin{tabular}{ccc}
\multicolumn{3}{c}{\small (a) Scissors} \\[2pt]
\subcaptionbox*{}{\includegraphics[height=2.5cm]{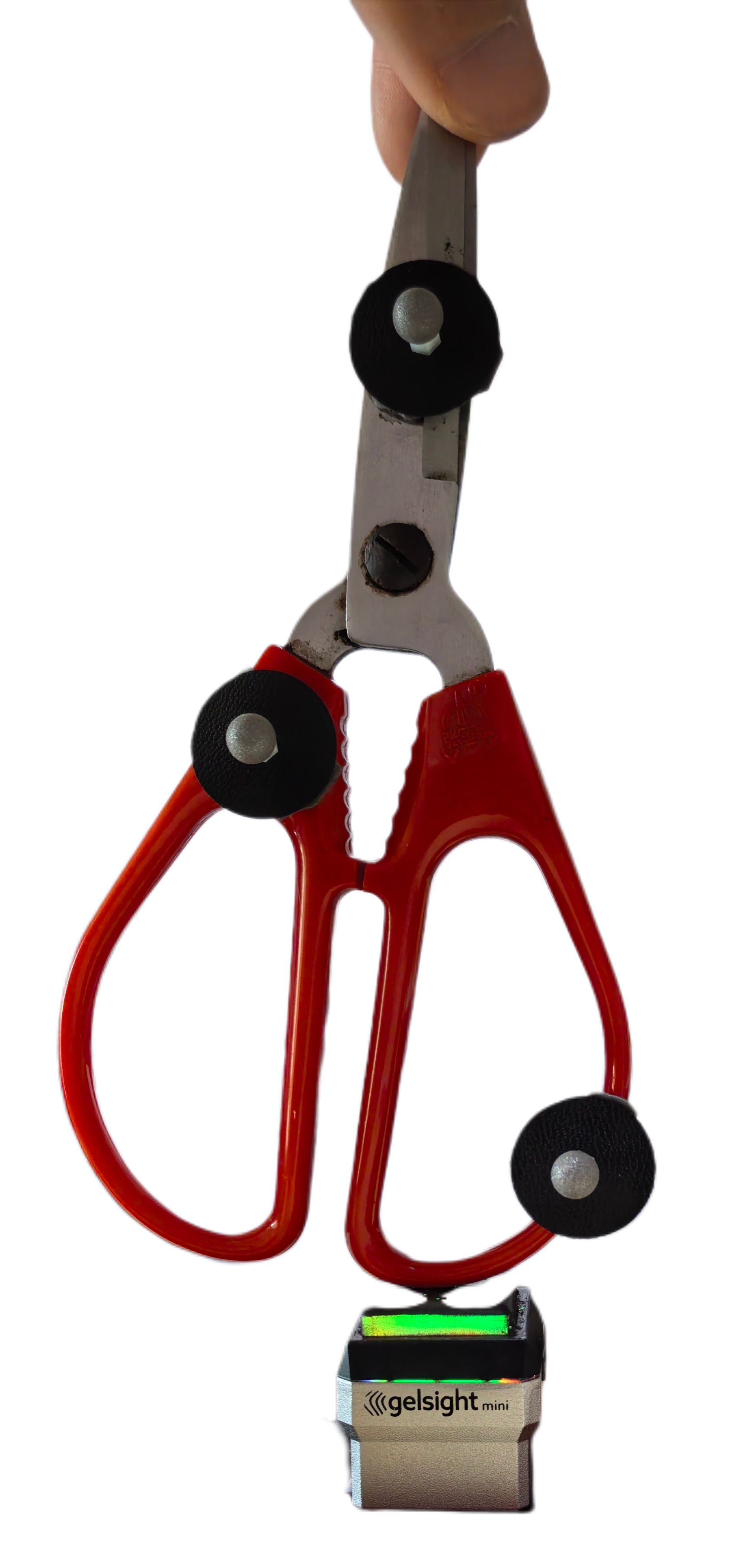}} &
\subcaptionbox*{}{\includegraphics[height=2.5cm]{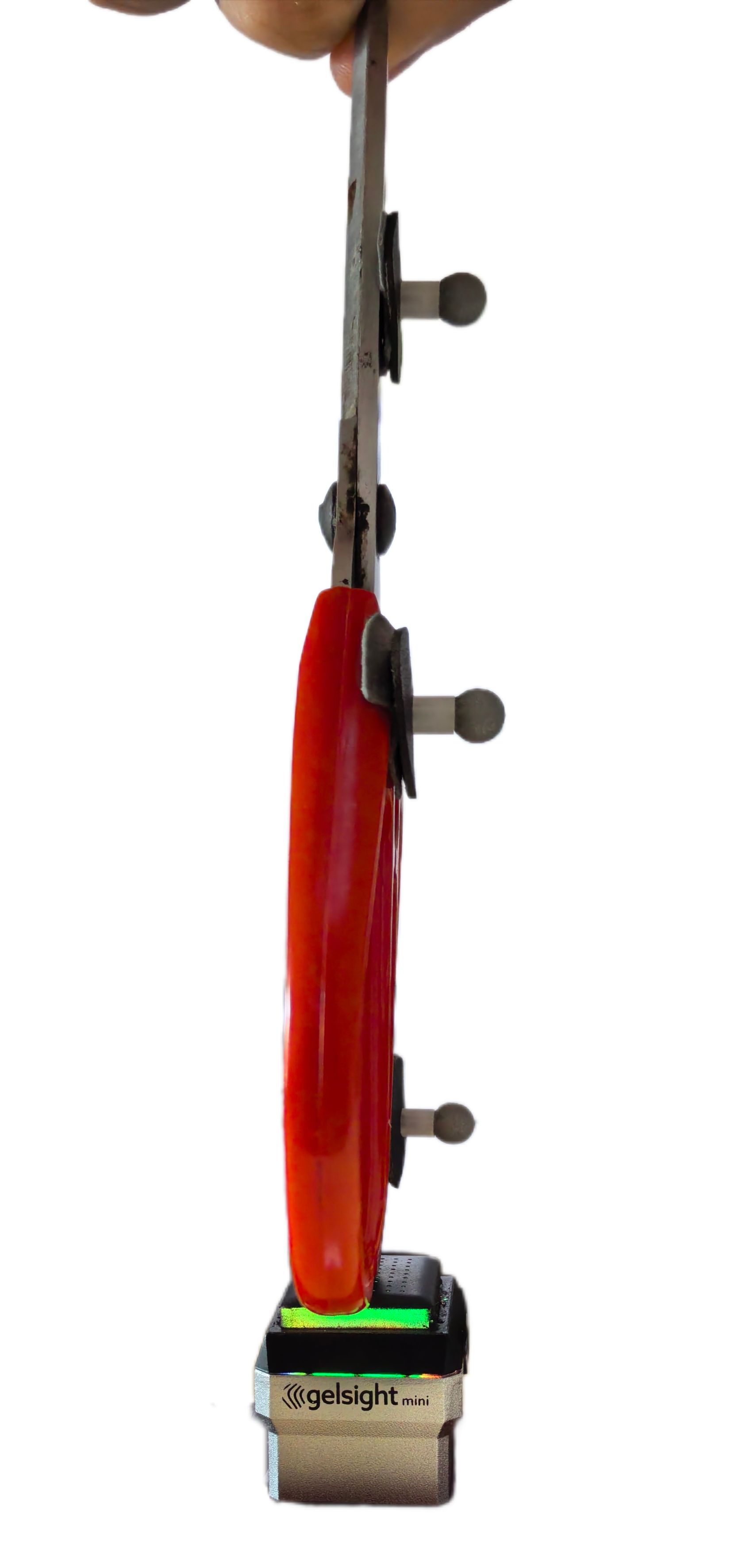}} &
\subcaptionbox*{}{\includegraphics[height=2.5cm]{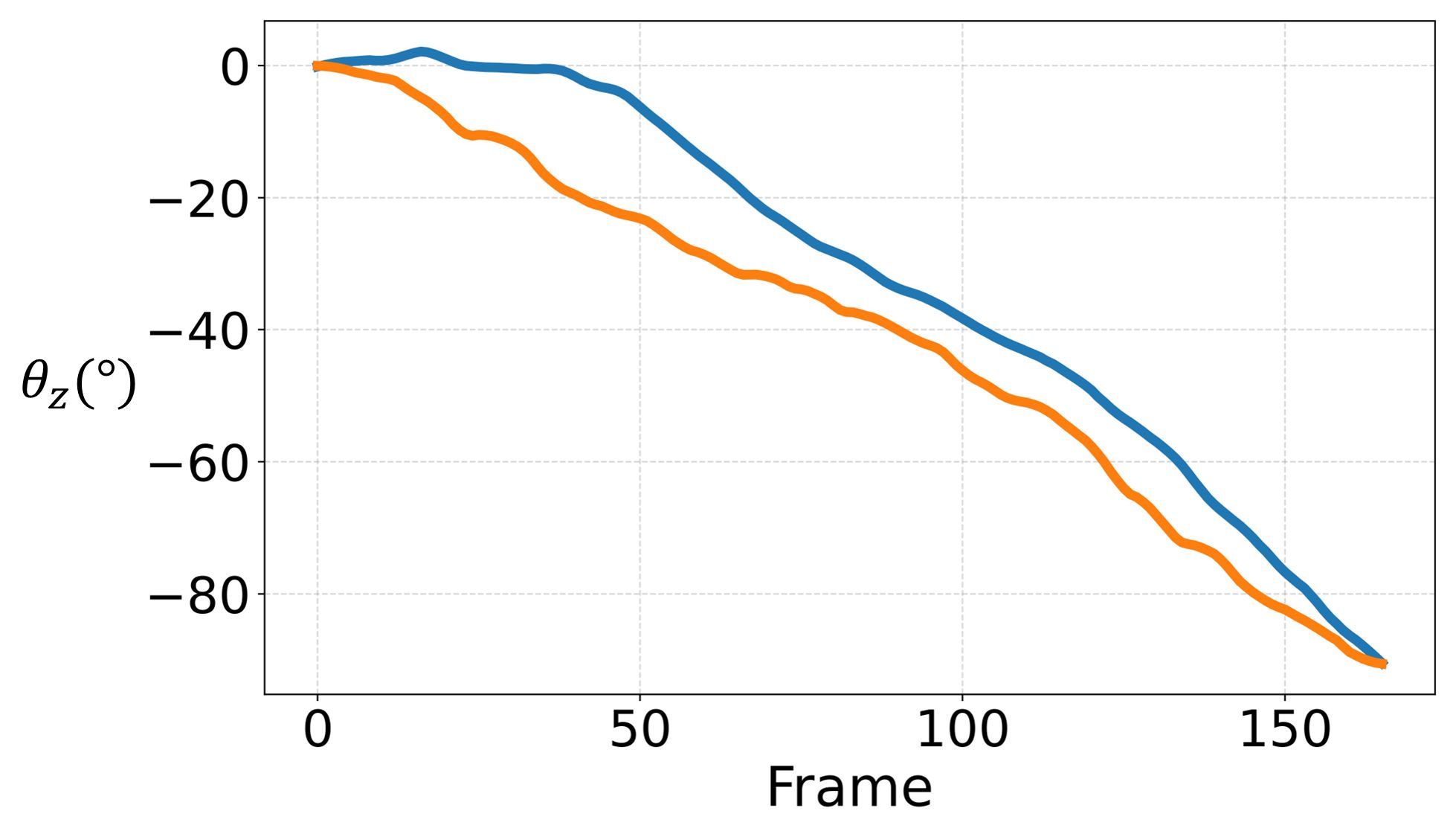}} \\[1pt]

\multicolumn{3}{c}{\small (b) Utility knife} \\[2pt]
\subcaptionbox*{}{\includegraphics[height=2.5cm]{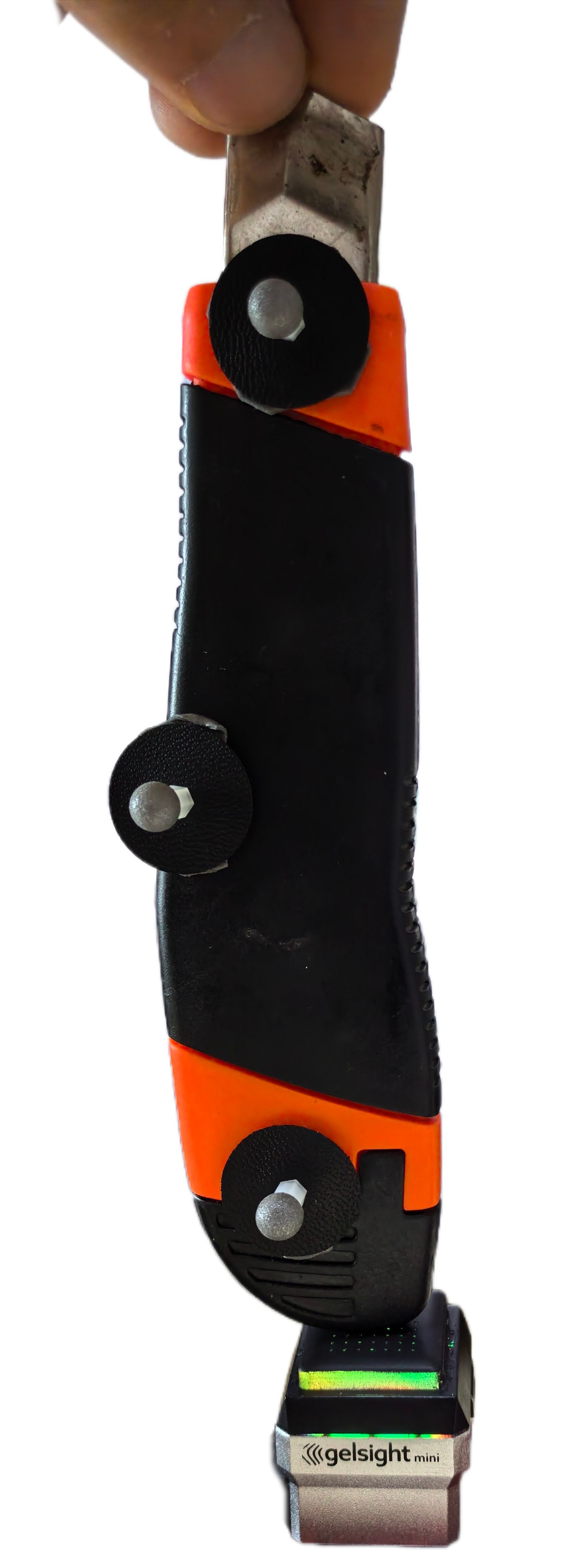}} &
\subcaptionbox*{}{\includegraphics[height=2.5cm]{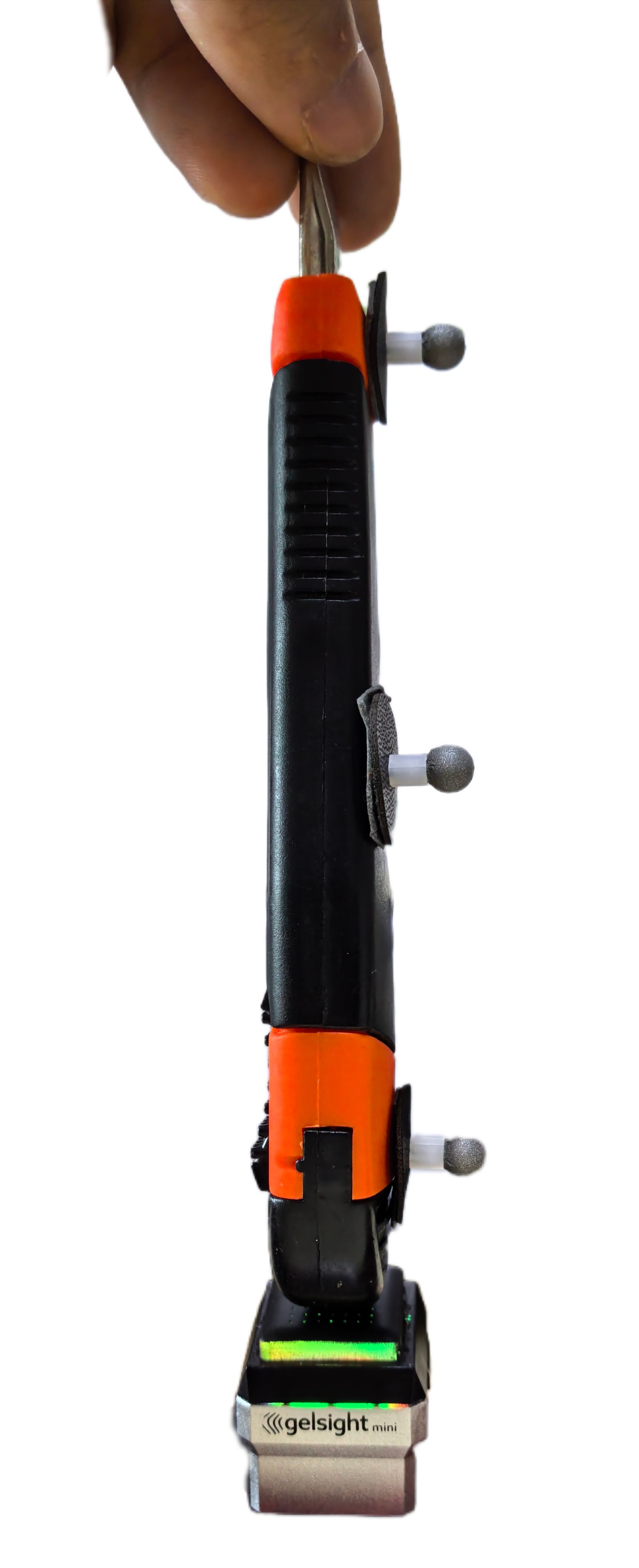}} &
\subcaptionbox*{}{\includegraphics[height=2.5cm]{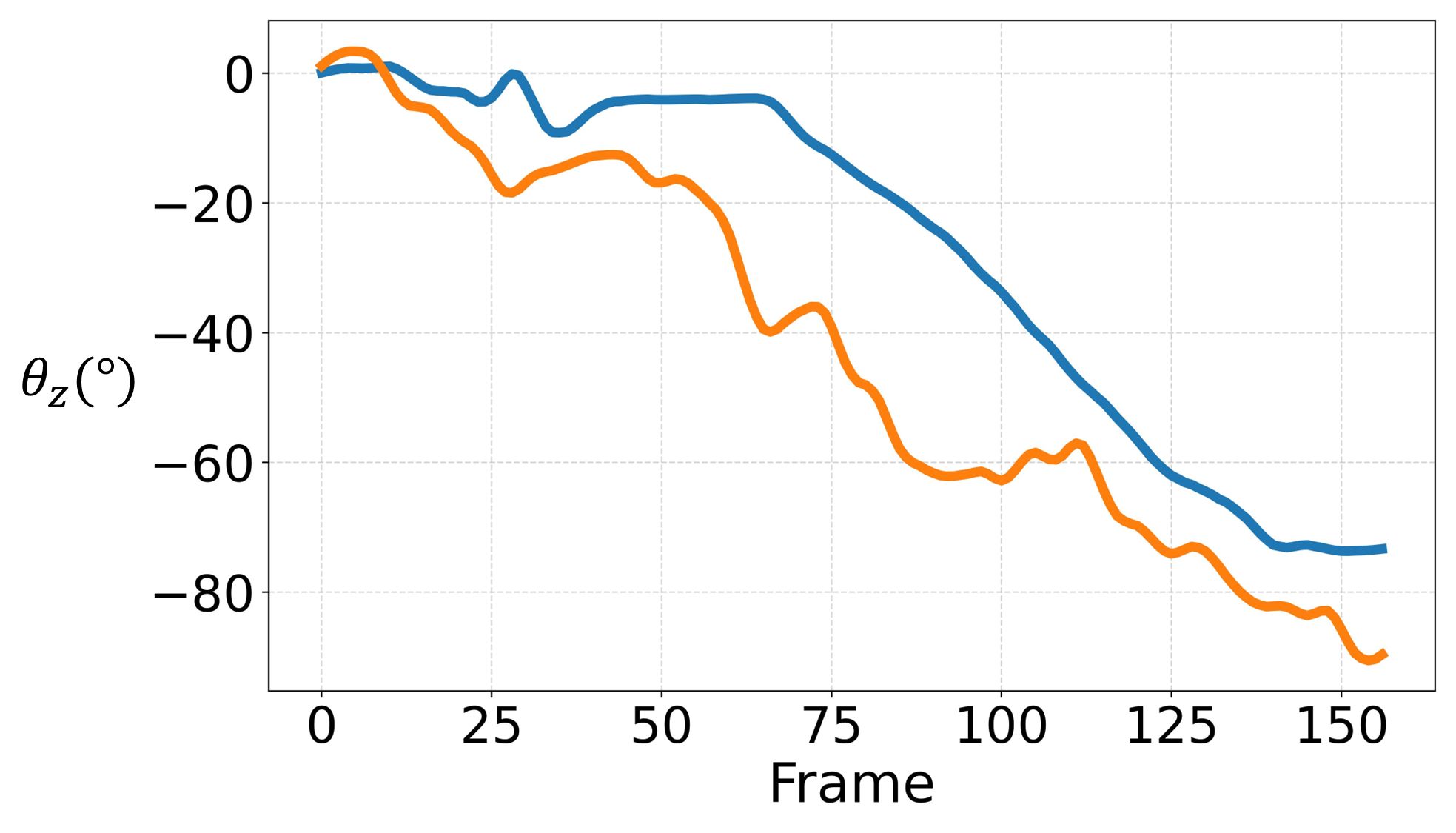}} \\[1pt]

\multicolumn{3}{c}{\small (c) Egg} \\[2pt]
\subcaptionbox*{}{\includegraphics[height=2.5cm]{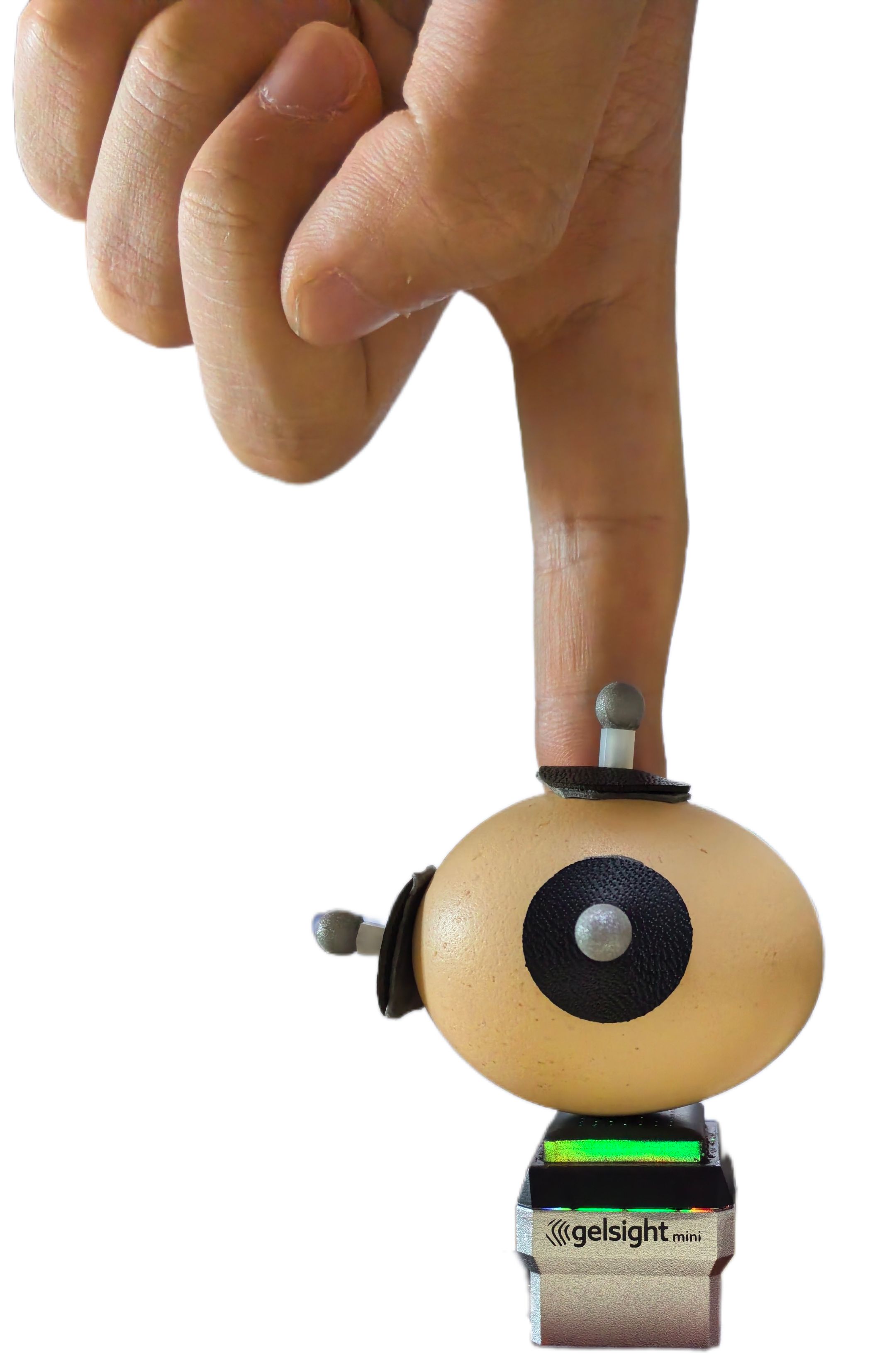}} &
\subcaptionbox*{}{\includegraphics[height=2.5cm]{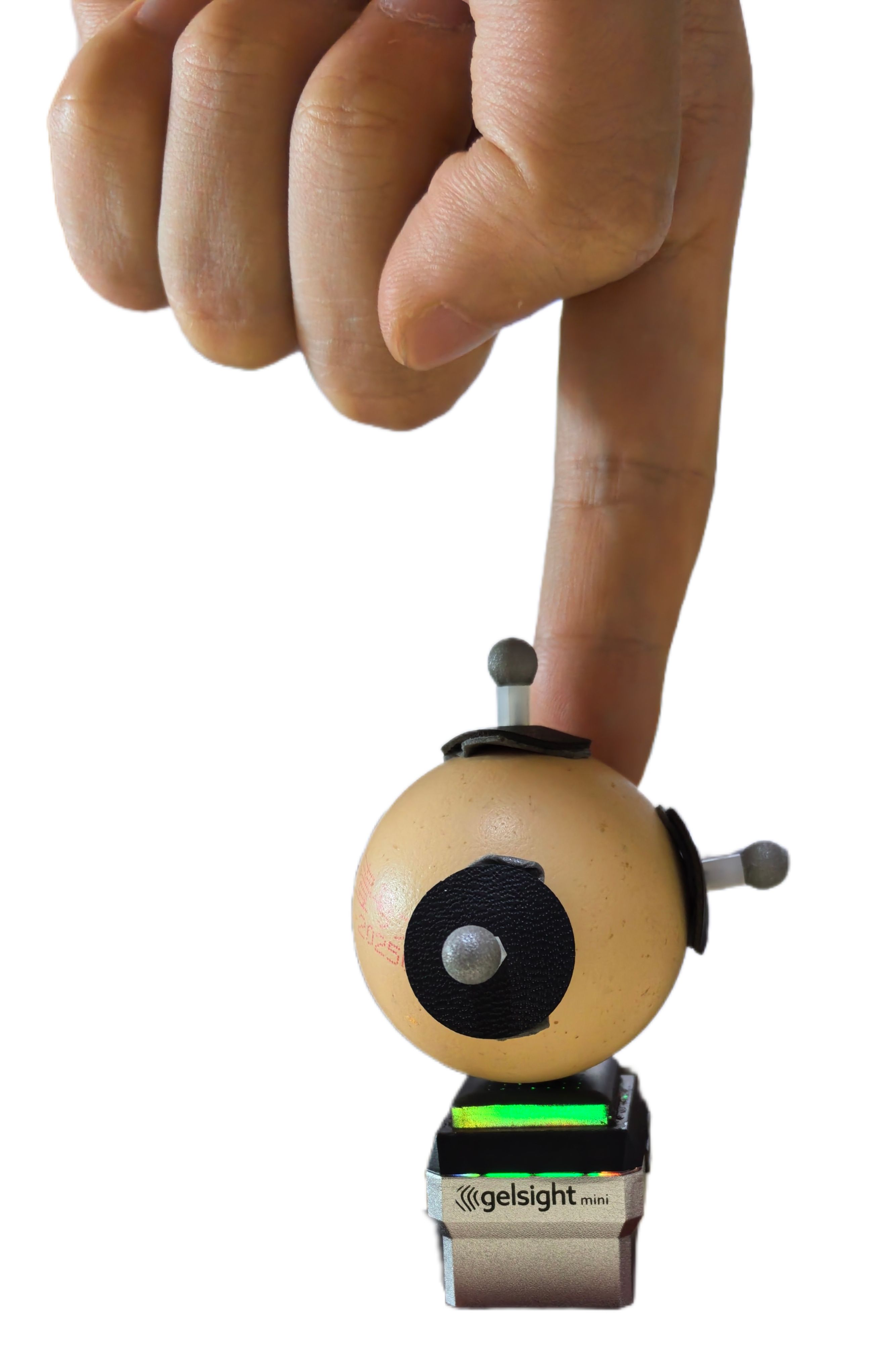}} &
\subcaptionbox*{}{\includegraphics[height=2.5cm]{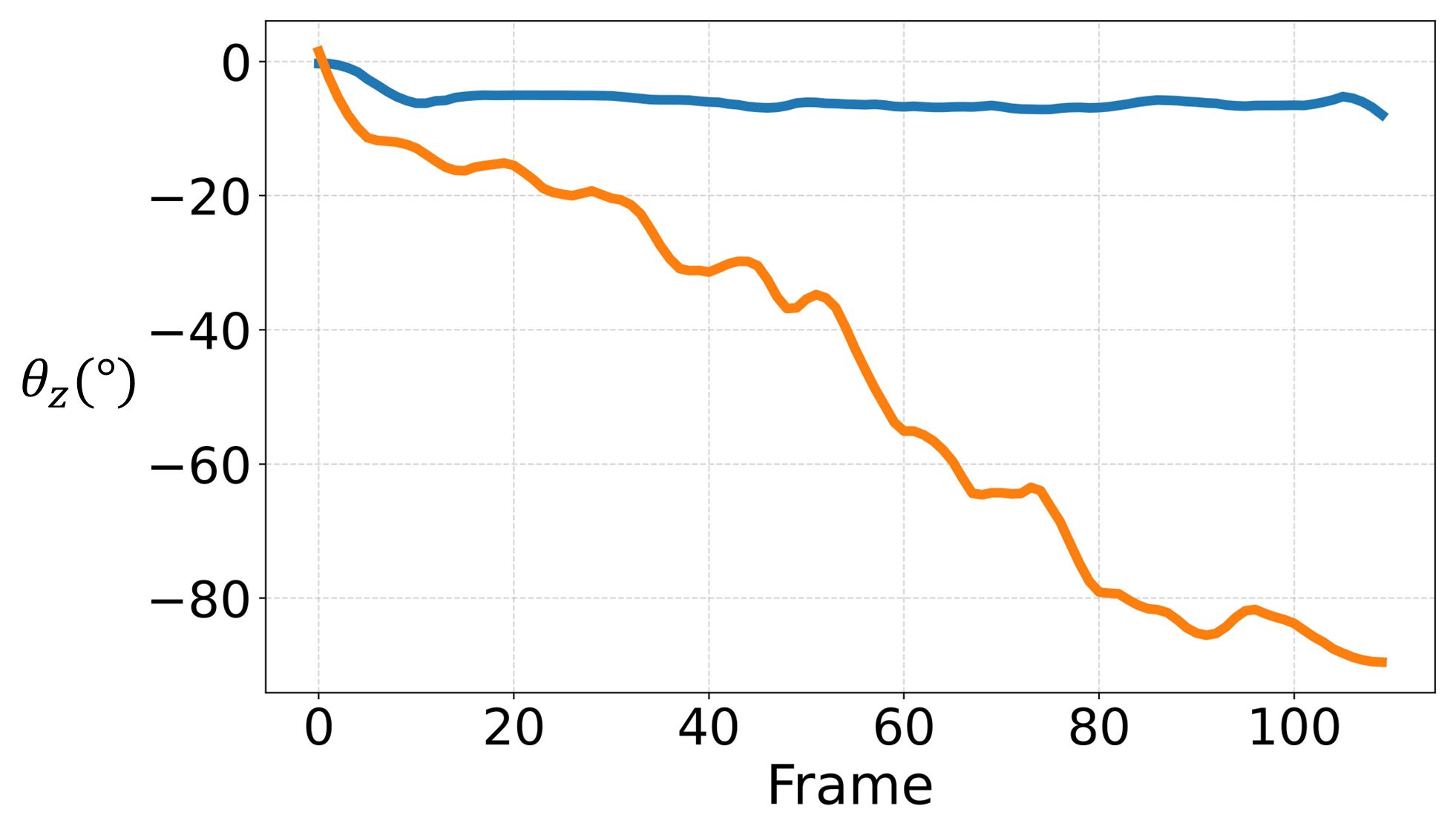}} \\[1pt]

\multicolumn{3}{c}{\small (d) Sensor box} \\[2pt]
\subcaptionbox*{}{\includegraphics[height=2.5cm]{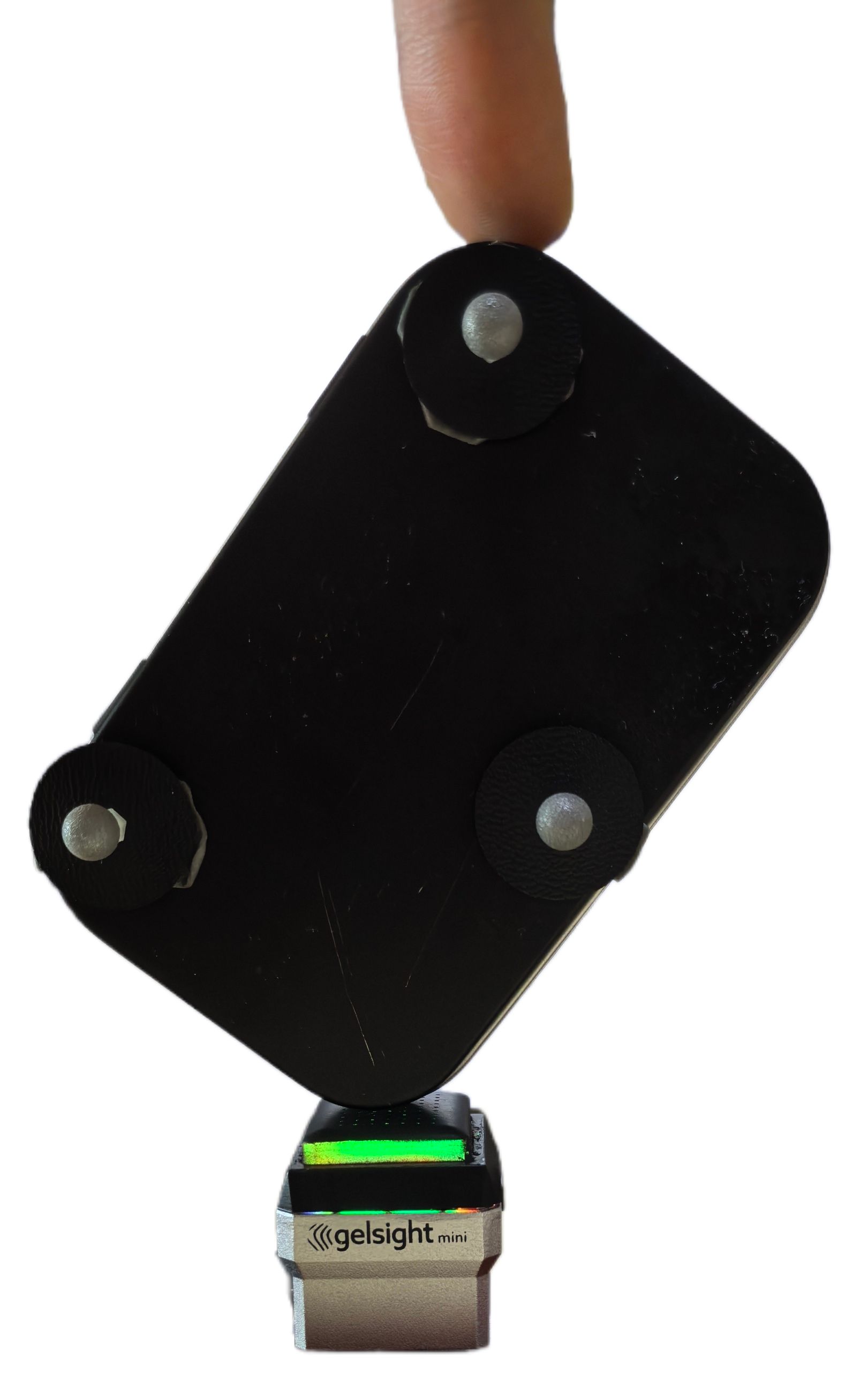}} &
\subcaptionbox*{}{\includegraphics[height=2.5cm]{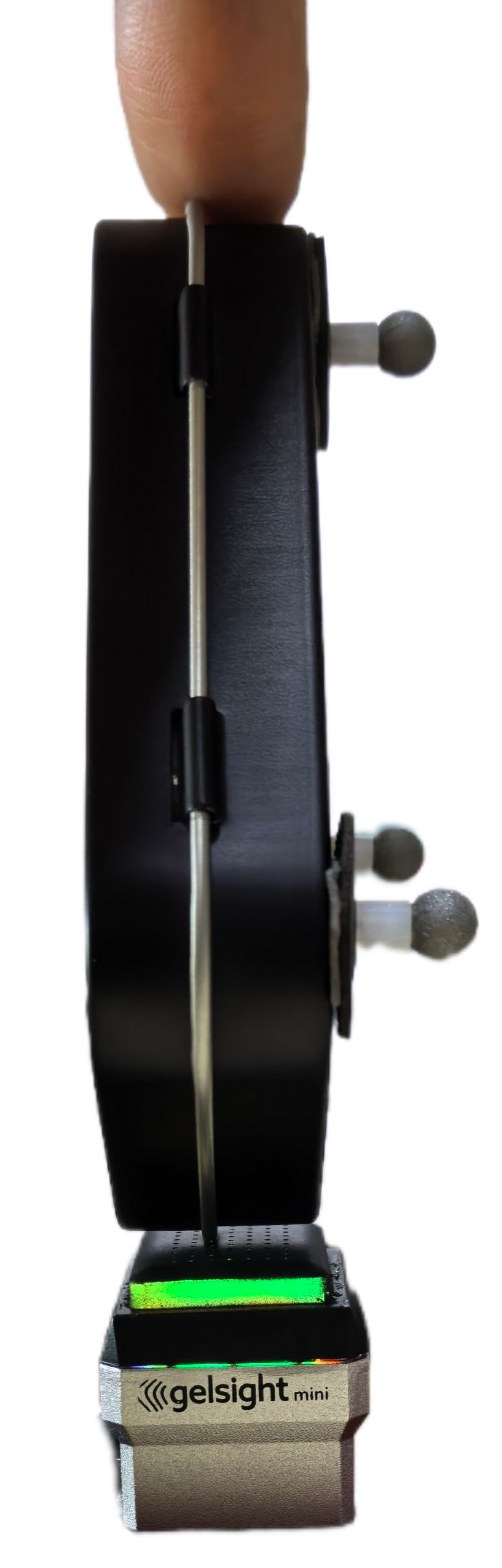}} &
\subcaptionbox*{}{\includegraphics[height=2.5cm]{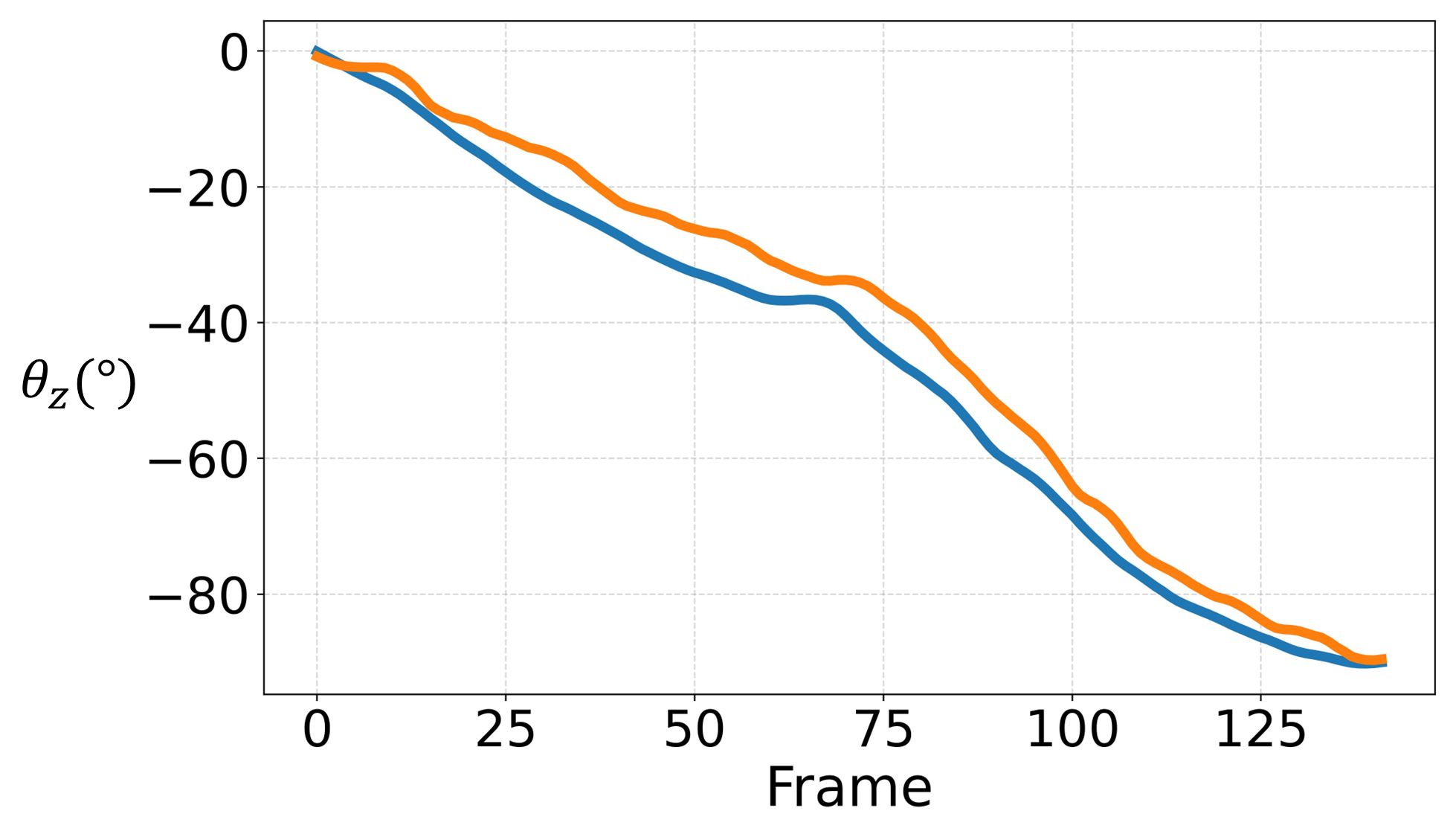}} \\
\end{tabular}}

\captionsetup{font=footnotesize}
\caption{Comparison of Z-axis rotation tracking for four objects. Each row shows the reference image and the estimated Z-axis rotation curves for both algorithms (Blue: NormalFlow; Yellow: InvariantCloud (Ours)), highlighting tracking performance across different object geometries. The vertical axis indicates the Z-axis rotation angle estimated by each algorithm. In each experiment, the object was controlled to rotate counterclockwise from 0° to -90°.}
\label{fig:ztrack}
\end{figure}

As shown by the angular tracking curves in Fig. 8, the NormalFlow method maintains relatively strong performance for Z-axis rotation tracking when objects possess distinct contour features, such as the handles of scissors or the edges of the sensor box. However, for objects with less distinctive surface contours—particularly the egg and the elliptical end of the utility knife—NormalFlow fails to track rotation effectively, with the estimated Z-axis angle remaining nearly static. In contrast, the proposed InvariantCloud method consistently achieves robust tracking even for such challenging objects, and further increasing the density of globally invariant point clouds can further enhance its performance.

\vspace{+5pt}
\setlength{\parindent}{0pt}
{\fontsize{10pt}{15pt}\selectfont \textit{D. Long-Term Tactile SLAM Registration Evaluation}}

\setlength{\parindent}{15pt}
To demonstrate the practical effectiveness of the inter-contact registration framework for real-world surface reconstruction tasks, an experiment was designed to challenge the system's ability to maintain consistent registration across multiple temporally separated contacts. This experiment is particularly important because vision-based tactile sensors like GelSight Mini have contact areas much smaller than typical object surfaces, requiring reliable integration of multiple contact patches to achieve complete reconstruction.

Surface contour reconstruction of the scissor object from the previous experiments was performed using only the GelSight Mini sensor, without any external motion capture devices or additional sensing modalities. This setup tests the algorithm's ability to rely solely on tactile information for maintaining geometric consistency across contacts. The experimental process and results are shown in Fig. 9, demonstrating successful reconstruction of the object's surface contour through multiple discrete contact events.

\begin{figure}[!tb]
    \centering
    \begin{subfigure}[b]{\linewidth}
        \centering
        \includegraphics[width=\linewidth]{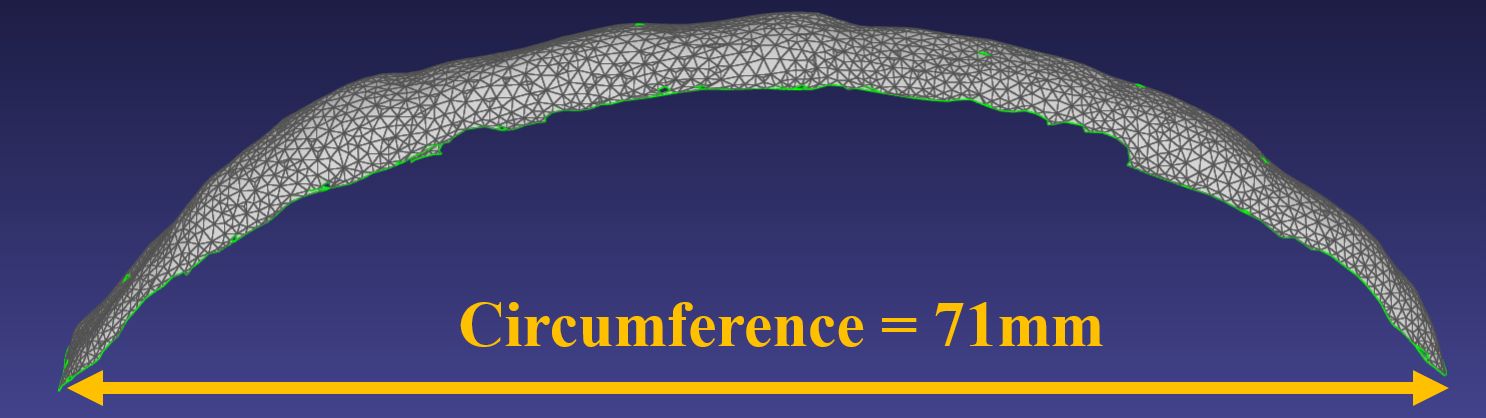}
        \captionsetup{font=footnotesize}
        \caption{Magnified view of the reconstructed surface model.}
        \label{fig:recon_a}
    \end{subfigure}
    \vspace{2mm}
    
    \begin{subfigure}[t]{0.48\linewidth}
        \centering
        \raisebox{2mm}{\includegraphics[width=\linewidth]{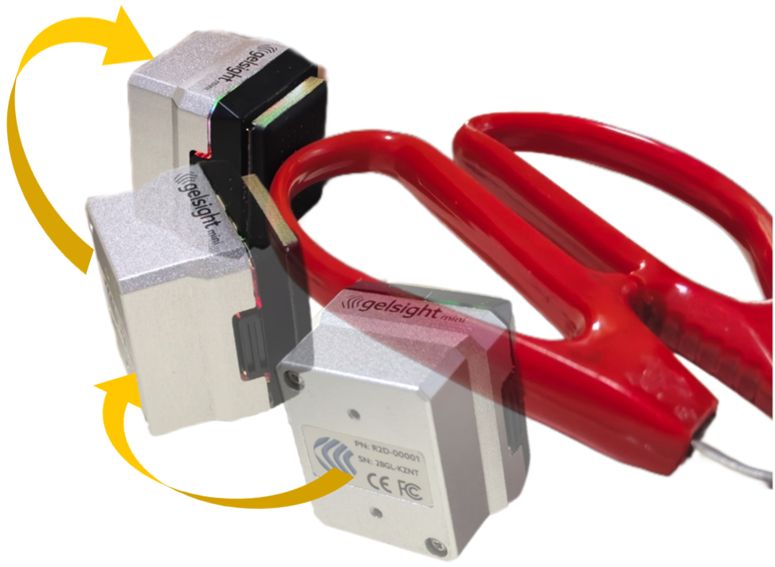}}
        \captionsetup{font=footnotesize}
        \caption{Sequential contact acquisition during surface reconstruction.}
        \label{fig:recon_c}
    \end{subfigure}
    \hfill
    \begin{subfigure}[t]{0.48\linewidth}
        \centering
        \includegraphics[width=\linewidth]{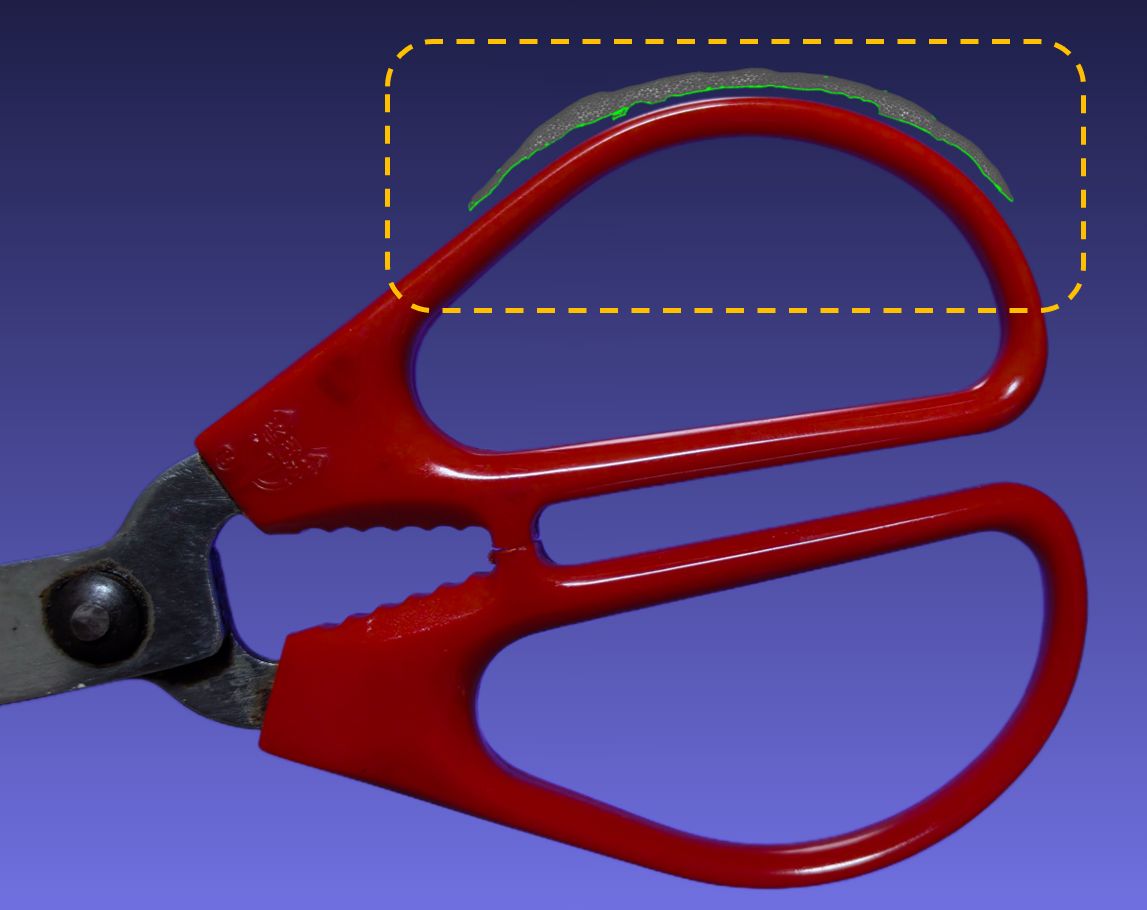}
        \captionsetup{font=footnotesize}
        \caption{Comparison between the reconstructed model and the physical object.}
        \label{fig:recon_b}
    \end{subfigure}
    
    \captionsetup{font=footnotesize}
    \caption{Multi-contact tactile SLAM: scanning workflow and surface reconstruction results.}
    \label{fig:recon_workflow}
\end{figure}

\section{Conclusion}
In this study, we propose InvariantCloud, a novel algorithm for 6D pose tracking of objects on visual-tactile sensor surfaces, based on globally invariant point clouds. By assigning unique IDs to each point, our method enables robust inter-frame correspondence and leverages the spatial distribution of these points to compute principal axes via PCA or efficiently solve for pose using the Kabsch algorithm. This approach achieves accurate and reliable pose estimation, outperforming existing baselines. However, a limitation remains when tracking perfectly spherical objects: the contact region remains nearly circular, making Z-axis rotation unobservable—a challenge for future research. With the reliable pose estimation provided by InvariantCloud, our next goal is to achieve complete tactile SLAM with tactile localization for common objects, enabling precise registration with 3D CAD models even in the absence of visual feedback. This advancement will allow accurate pose estimation based on local contact images and global models, removing the constraint of continuous sensor contact, and opening new possibilities for high-precision manipulation and control tasks, including contact-force-based closed-loop manufacturing processes \cite{Yang2024ContactForce}.

\bibliographystyle{IEEEtran}
\bibliography{mybib}

\end{document}